# Sparse Nonlinear Regression: Parameter Estimation and Asymptotic Inference

Zhuoran Yang[*]   Zhaoran Wang[*]   Han Liu[*]

Yonina C. Eldar [†]   Tong Zhang [‡]

November 11, 2015

## Abstract

We study parameter estimation and asymptotic inference for sparse nonlinear regression. More specifically, we assume the data are given by $y = f(\mathbf{x}^\top \boldsymbol{\beta}^*) + \epsilon$, where $f$ is nonlinear. To recover $\boldsymbol{\beta}^*$, we propose an $\ell_1$-regularized least-squares estimator. Unlike classical linear regression, the corresponding optimization problem is nonconvex because of the nonlinearity of $f$. In spite of the nonconvexity, we prove that under mild conditions, every stationary point of the objective enjoys an optimal statistical rate of convergence. In addition, we provide an efficient algorithm that provably converges to a stationary point. We also access the uncertainty of the obtained estimator. Specifically, based on any stationary point of the objective, we construct valid hypothesis tests and confidence intervals for the low dimensional components of the high-dimensional parameter $\boldsymbol{\beta}^*$. Detailed numerical results are provided to back up our theory.

## 1 Introduction

We study a family of sparse nonlinear regression models. Let $\boldsymbol{\beta}^* = (\beta_1^*, \ldots, \beta_d^*)^\top \in \mathbb{R}^d$ be the sparse parameter vector of interest. We consider the model

$$y = f(\mathbf{x}^\top \boldsymbol{\beta}^*) + \epsilon, \qquad (1.1)$$

where $y \in \mathbb{R}$ is a response variable, $\mathbf{x} \in \mathbb{R}^d$ is the covariate and $\epsilon \in \mathbb{R}$ is the exogenous noise. When $f$ is the identity function, (1.1) reduces to the well studied linear model. Given independent and identically distributed observations $\{y_i, \mathbf{x}_i\}_{i=1}^n$, our goal is to estimate $\boldsymbol{\beta}^*$ even when $d \gg n$.

We can view (1.1) as a perceptron with noise, which is the basic building block of a feed forward neural network (Rumelhart et al., 1986). Establishing the theoretical guarantees of the estimation in (1.1) may provide insight on more complicated neural networks. Our model is also inspired by the nonlinear sparse recovery problems (Beck and Eldar, 2013a,b; Aksoylar and Saligrama, 2014) which aim to recover a sparse parameter from a nonlinear system.

---

[*]Department of Operations Research and Financial Engineering, Princeton University, Princeton, NJ 08544, USA; e-mail: {zy6,zhaoran,hanliu}@princeton.edu.

[†]Department of EE Technion, Israel Institute of Technology, Haifa 32000, Israel; e-mail: yonina@ee.technion.ac.il.

[‡]Department of Statistics, Rutgers University, Piscataway, New Jersey 08854, USA; e-mail: tzhang@stat.rutgers.edu

## 1.1 Main Results

Assuming $f$ is monotonic, a straightforward way to estimate $\boldsymbol{\beta}^*$ is to solve a sparse linear regression problem (Eldar and Kutyniok, 2012) using the transformed data $\{f^{-1}(y_i), \mathbf{x}_i\}_{i=1}^n$. However, this approach works well only in the noiseless case with $\epsilon = 0$. Otherwise, it results in inaccurate parameter estimation and high prediction error due to the inverse operation. In this paper, we propose estimating the parameter $\boldsymbol{\beta}^*$ by solving the following $\ell_1$-regularized least-squares problem:

$$\underset{\boldsymbol{\beta} \in \mathbb{R}^d}{\text{minimize}} \, \frac{1}{n} \sum_{i=1}^n \big[y_i - f(\mathbf{x}_i^\top \boldsymbol{\beta})\big]^2 + \lambda \|\boldsymbol{\beta}\|_1, \tag{1.2}$$

where $\lambda$ is a regularization parameter and $\|\cdot\|_1$ is the vector $\ell_1$-norm. Unlike the linear model for which (1.2) is a convex optimization problem, in general settings (1.2) could be highly nonconvex due to the nonlinearity of $f$, which prevents us from obtaining the global optimum. The existence of $f$ also prevents us from having the restricted strongly convex property of the loss function.

In spite of the challenge of nonconvexity, we prove that any stationary point $\widehat{\boldsymbol{\beta}}$ of (1.2) enjoys optimal statistical rates of convergence under suitable conditions, i.e., with high probability

$$\big\|\widehat{\boldsymbol{\beta}} - \boldsymbol{\beta}^*\big\|_2 \leq C_1 \cdot \sqrt{s^* \log d / n} \quad \text{and} \quad \big\|\widehat{\boldsymbol{\beta}} - \boldsymbol{\beta}^*\big\|_1 \leq C_2 \cdot s^* \sqrt{\log d / n},$$

where $s^*$ is the number of nonzero entries of $\boldsymbol{\beta}^*$ and $C_1$, $C_2$ are some absolute constants which do not depend on $n$, $d$ or $s^*$. The statistical rates of convergence cannot be improved even when $f$ is the identity function. In addition, we require a scaling of $n = O(s^* \log d)$ samples to obtain a vanishing error, which is also needed for linear sparse recovery problems (Eldar and Kutyniok, 2012). Next, we provide an efficient gradient-based algorithm that provably converges to a stationary point. Our method is iterative and consist of soft-thresholding after a gradient descent step. This approach can be viewed as a generalization of the ISTA algorithm (Beck and Teboulle, 2009) to the nonlinear setting.

Once an estimator $\widehat{\boldsymbol{\beta}}$ is obtained, another important task is to assess the uncertainty of the estimator. Specifically, let $\widehat{\beta}_j$ be the $j$th entry of $\widehat{\boldsymbol{\beta}}$. We aim to test the hypothesis $H_0 : \beta_j^* = 0$. For this, we propose methods to construct confidence intervals as well as hypothesis tests for low-dimensional components of the high-dimensional parameter $\boldsymbol{\beta}^*$. Theoretical guarantees on these inferential methods are also provided.

## 1.2 Related Work

The model in (1.1) is closely related to the single index model, which assumes $(y, \mathbf{x})$ satisfy $y = f(\mathbf{x}^\top \boldsymbol{\beta}^*) + \epsilon$ with an unknown $f$. The single index model is well studied in low dimensional settings where $d \ll n$. See, e.g., McCullagh et al. (1989); Horowitz (2000); Härdle et al. (1993); Ichimura (1993); Sherman (1994); Xia and Li (1999); Xia et al. (1999); Delecroix et al. (2000, 2006) and references therein. They mostly consider $M$-estimators that simultaneously estimate $f$ and $\boldsymbol{\beta}^*$. However, these $M$-estimators are defined as the global optima of nonconvex minimization problems which are intractable to obtain. In high-dimensional settings where $\boldsymbol{\beta}^*$ is sparse, Alquier and Biau (2013) establish PAC-Bayesian analysis for sparse single index models. Plan et al. (2014); Plan and Vershynin (2015) propose marginal regression and generalized Lasso estimators which attain fast statistical rates of convergence. Nevertheless, the flexibility of the unknown link function $f$ comes at a price. In detail, Plan et al. (2014); Plan and Vershynin (2015) require $\mathbf{x}$ to be exactly Gaussian for their methods to succeed, even if $f$ is known a priori. Also, unknown $f$ raises identifiability issues, since the magnitude of $\boldsymbol{\beta}^*$ can be incorporated into $f$. As a result, these methods only estimate the direction of $\boldsymbol{\beta}^*$.

Another related line of work is sufficient dimension reduction, for which we aim to recover a subspace $\mathcal{U}$ such that $y$ only depends on the projection of $\mathbf{x}$ onto $\mathcal{U}$. Both single index model and our problem can be viewed as special cases of the framework in which $\mathcal{U}$ is a one-dimensional subspace. See Li (1991, 1992); Cook (1998); Cook and Lee (1999); Li (2007) and the references therein. Most works in this direction use spectral methods, which also rely on the Gaussian assumption and can only estimate the direction of $\boldsymbol{\beta}^*$. In comparison, we assume $f$ is known. In this setting, we allow $\mathbf{x}$ to follow more general distributions and can directly estimate $\boldsymbol{\beta}^*$. Kalai and Sastry (2009); Kakade et al. (2011) propose iterative algorithms that alternatively estimate $f$ and $\boldsymbol{\beta}^*$ based on the isotonic regression in the setting with $d \ll n$. However, their analysis focuses on generalization error instead of estimation error, which is the primary goal in this paper.

Our work is also related to problems of phase retrieval where the goal is to recover a signal $\boldsymbol{\beta}^* \in \mathbb{C}^d$ from the magnitude of its linear measurements contaminated by random noise. More specifically, the model of phase retrieval is given by $y = |\mathbf{x}^\top \boldsymbol{\beta}|^2 + \epsilon$. For high-dimensional settings, this problem is extensively studied under noisy or noiseless settings. See, e.g., Jaganathan et al. (2012); Ohlsson et al. (2012); Li and Voroninski (2013); Candès et al. (2013); Eldar and Mendelson (2014); Shechtman et al. (2014, 2015); Ohlsson and Eldar (2014); Candès et al. (2015); Waldspurger et al. (2015); Eldar et al. (2015); Cai et al. (2015); Tu et al. (2015). These works show that a high dimensional signal can be accurately estimated up to global phase under restrictive assumptions on $\mathbf{x}$, e.g., $\mathbf{x}$ is Gaussian or certain classes of measurements. However, our work considers general measurements. Note that phase retrieval does not fall in the model under (1.1) because it uses a quadratic function, which is not monotonic. See §4 for a more detailed discussion.

In terms of asymptotic inference, there is a recent surge of research focusing on high dimensional inference for generalized linear models (Meinshausen et al., 2009; Wasserman and Roeder, 2009; Meinshausen and Bühlmann, 2010; Lockhart et al., 2014; Taylor et al., 2014; Zhang and Zhang, 2014; Javanmard and Montanari, 2013; van de Geer et al., 2014). Most of these works rely on the fact that the estimators are the global optima of convex optimization problems, which is not the case in our setting. For estimation and inference in the presence of nonconvexity, Zhang and Zhang (2012); Loh and Wainwright (2014); Wang et al. (2014); Loh (2015) establish both computational and statistical theory for nonconvex $M$-estimators when the restricted strongly convex property of the loss function is satisfied.

## 1.3 Main Contribution

Our contribution is twofold. First, we propose an $\ell_1$-regularized least-squares estimator for parameter estimation. We prove that every stationary point of (1.2) converges to the true parameter, which explains the empirical success of regularized least-squares in the presence of nonlinear transforms. In the noiseless setting, as long as the number of samples is proportional to $s^* \log d$, we are able to exactly recover $\boldsymbol{\beta}^*$. Moreover, the proposed algorithm is shown to be minimax optimal. To the best of our knowledge, this is the first parameter estimation result for the model (1.1) in high dimensional settings that does not rely on the normality of $\mathbf{x}$, and recovers both the magnitude and direction of $\boldsymbol{\beta}^*$. Our analysis for the stationary points of nonconvex optimization problems is of independent interest. Second, we provide statistical inferential procedures to assess the uncertainty associated with these stationary points. Specifically, we construct the hypothesis tests and confidence intervals for the low-dimensional components of the high-dimensional sparse parameter. The type-I errors of the hypothesis tests and the coverage probabilities of the confidence intervals are shown to converge to the nominal significance level asymptotically.

**Organization of the rest of this paper** In §2 we present our method for parameter estimation and asymptotic inference. We lay out the theory in §3 and provide a proof of the main results in §6. Connection to prior work is discussed in §4. We corroborate our theoretical results with thorough numerical results in §5. In §7 we conclude the paper.

## 2 High-dimensional Estimation and Statistical Inference

In this section, we introduce the proposed methods for parameter estimation and asymptotic inference. In addition, we present the intuition behind our methods and compare our estimation procedures with the one that inverts the nonlinear function $f$ directly.

### 2.1 Parameter Estimation

Recall that we observe $\{(y_i, \mathbf{x}_i)\}_{i=1}^n$ satisfying $y_i = f(\mathbf{x}_i^\top \boldsymbol{\beta}^*) + \epsilon_i$. We assume the function $f$ is monotonic and continuously differentiable. We define the least-square loss function as

$$L(\boldsymbol{\beta}) = \frac{1}{2n} \sum_{i=1}^n [y_i - f(\mathbf{x}_i^\top \boldsymbol{\beta})]^2. \tag{2.1}$$

We assume $\boldsymbol{\beta}^*$ is sparse and estimate it by solving the $\ell_1$-regularized optimization problem in (1.2).

Due to the nonlinearity of $f$, $L(\boldsymbol{\beta})$ can be nonconvex. As a result, we can only find a stationary point $\widehat{\boldsymbol{\beta}}$ satisfying $\nabla L(\widehat{\boldsymbol{\beta}}) + \lambda \cdot \boldsymbol{\xi} = \mathbf{0}$, where $\boldsymbol{\xi} \in \partial \|\widehat{\boldsymbol{\beta}}\|_1$ and $\nabla L(\boldsymbol{\beta})$ is the gradient of $L(\boldsymbol{\beta})$. To obtain a stationary point, we apply the proximal gradient method, which generates an iterative sequence $\{\boldsymbol{\beta}^{(t)}, t \geq 0\}$ satisfying

$$\boldsymbol{\beta}^{(t+1)} = \underset{\boldsymbol{\beta} \in \mathbb{R}^d}{\operatorname{argmin}} \{\langle \nabla L(\boldsymbol{\beta}^{(t)}), \boldsymbol{\beta} - \boldsymbol{\beta}^{(t)} \rangle + \alpha_t/2 \cdot \|\boldsymbol{\beta} - \boldsymbol{\beta}^{(t)}\|_2^2 + \lambda \|\boldsymbol{\beta}\|_1\}, \tag{2.2}$$

where $1/\alpha_t > 0$ is the stepsize at the $t$-th iteration. In our setting, $\nabla L(\boldsymbol{\beta}^{(t)})$ is given by

$$\nabla L(\boldsymbol{\beta}^{(t)}) = -\frac{1}{n} \sum_{i=1}^n [y_i - f(\mathbf{x}_i^\top \boldsymbol{\beta}^{(t)})] f'(\mathbf{x}_i^\top \boldsymbol{\beta}^{(t)}) \mathbf{x}_i. \tag{2.3}$$

Solving (2.2) with $\nabla L(\boldsymbol{\beta}^{(t)})$ given in (2.3) results in

$$\boldsymbol{\beta}^{(t+1)} = \underset{\boldsymbol{\beta} \in \mathbb{R}^d}{\operatorname{argmin}} \{1/2 \cdot \|\boldsymbol{\beta} - \mathbf{u}^{(t)}\|_2^2 + \lambda/\alpha_t \cdot \|\boldsymbol{\beta}\|_1\},$$

where $\mathbf{u}^{(t)}$ is given by $\mathbf{u}^{(t)} := \boldsymbol{\beta}^{(t)} - 1/\alpha_t \cdot \nabla L(\boldsymbol{\beta}^{(t)})$. This problem has an explicit solution given by

$$\beta_i^{(t+1)} = \operatorname{soft}(u_i^{(t)}, \lambda/\alpha_t) \quad \text{for } 1 \leq i \leq d, \tag{2.4}$$

where $\operatorname{soft}(u, a) := \operatorname{sign}(u) \max\{|u| - a, 0\}$ is the soft-thresholding operator.

The resulting algorithm is given in Algorithm 1, which is an application of the SpaRSA method proposed by Wright et al. (2009) to our nonconvex problem. The main step is given in (2.4), which performs a soft-thresholding step on a gradient-descent update. This algorithm reduces to ISTA (Beck and Teboulle, 2009) when $f$ is the identity. For nonlinear sparse recovery problems, this technique is also similar to the thresholded Wirtinger flow algorithm proposed for phase retrieval (Candès et al., 2015; Cai et al., 2015).

To pick a suitable $\alpha_t$, we use the line search procedure described in Algorithm 2. It iteratively increases $\alpha_t$ by a factor of $\eta$ to ensure that $\boldsymbol{\beta}^{(t+1)}$ satisfies the acceptance criterion, which guarantees sufficient decrease of the objective function. To choose the initial $\alpha_t$ at the beginning of each line search iteration, we use the Barzilai-Borwein (BB) spectral method (Barzilai and Borwein, 1988) in Algorithm 2, which guarantees that the initial value of each stepsize $\alpha_t$ lies in the interval $[\alpha_{\min}, \alpha_{\max}]$. Using the theory of Wright et al. (2009), we establish the numerical convergence of the iterative sequence to a stationary point of (1.2) in §6. However, it is challenging to establish the statistical properties of the stationary points. Our theory in §3 shows that, surprisingly, any stationary point enjoys satisfactory statistical guarantees. Consequently, Algorithm 1 yields a stationary point that is desired for parameter estimation.

**Algorithm 1** Proximal gradient algorithm for solving the $\ell_1$-regularized problem in (1.2).

1: **Input:** regularization parameter $\lambda > 0$, update factor $\eta > 1$, constants $\zeta > 0$, $\alpha_{\min}, \alpha_{\max}$ with $0 < \alpha_{\min} < 1 < \alpha_{\max}$, integer $M > 0$, and $\phi(\boldsymbol{\beta}) := L(\boldsymbol{\beta}) + \lambda \|\boldsymbol{\beta}\|_1$
2: **Initialization:** set the iteration counter $t \leftarrow 0$ and choose $\boldsymbol{\beta}^{(0)} \in \mathbb{R}^d$
3: **Repeat**
4:     Choose stepsize $\alpha_t$ according to Algorithm 2
5:     **Repeat**
6:         $\mathbf{u}^{(t)} \leftarrow \boldsymbol{\beta}^{(t)} + \frac{1}{n\alpha_t} \cdot \sum_{i=1}^n \left[y_i - f(\mathbf{x}_i^\top \boldsymbol{\beta}^{(t)})\right] f'(\mathbf{x}_i^\top \boldsymbol{\beta}^{(t)}) \mathbf{x}_i.$
7:         $\beta_i^{(t+1)} \leftarrow \mathrm{soft}(u_i^{(t)}, \lambda/\alpha_t)$ for $1 \leq i \leq d$.
8:         $\alpha_t \leftarrow \eta \cdot \alpha_t$
9:     **Until** $\boldsymbol{\beta}^{(t+1)}$ satisfies the acceptance criterion:
10:        $\phi(\boldsymbol{\beta}^{(t+1)}) \leq \max\{\phi(\boldsymbol{\beta}^{(j)}) - \zeta \cdot \alpha_t/2 \cdot \|\boldsymbol{\beta}^{(t+1)} - \boldsymbol{\beta}^{(t)}\|_2^2 : \max(t-M, 0) \leq j \leq t\}$
11:     Update the iteration counter $t \leftarrow t + 1$
12: **Until** $\|\boldsymbol{\beta}^{(t)} - \boldsymbol{\beta}^{(t-1)}\|_2 / \|\boldsymbol{\beta}^{(t)}\|_2$ is sufficiently small
13: **Output:** $\widehat{\boldsymbol{\beta}} \leftarrow \boldsymbol{\beta}^{(t)}$

---

**Algorithm 2** The Barzilai-Borwein (BB) spectral approach for choosing $\alpha_t$ in Line 4 of Algorithm 1.

1: **Input:** the iteration counter $t$, $\boldsymbol{\delta}^{(t)} = \boldsymbol{\beta}^{(t)} - \boldsymbol{\beta}^{(t-1)}$ and $\boldsymbol{g}^{(t)} = \nabla L(\boldsymbol{\beta}^{(t)}) - \nabla L(\boldsymbol{\beta}^{(t-1)})$
2: **if** $t = 0$ **then**
3:     **Output:** $\alpha_t = 1$
4: **else**
5:     **Output:** $\alpha_t = \langle \boldsymbol{\delta}^{(t)}, \boldsymbol{g}^{(t)} \rangle / \langle \boldsymbol{\delta}^{(t)}, \boldsymbol{\delta}^{(t)} \rangle$ or $\alpha_t = \langle \boldsymbol{g}^{(t)}, \boldsymbol{g}^{(t)} \rangle / \langle \boldsymbol{\delta}^{(t)}, \boldsymbol{g}^{(t)} \rangle$
6: **end if**

---

When $f$ is known, it seems tempting to apply linear compressed sensing procedures to the inverted data $\{z_i, \mathbf{x}_i\}$ where $z_i = f^{-1}(y_i)$. If $f$ is linear, say $f(u) = au + b$, then $f^{-1}(u) = a^{-1}(u-b)$. In this case we have $z = f^{-1}(y) = \mathbf{x}^\top \boldsymbol{\beta}^* + a^{-1}\epsilon$, which is exactly a linear model. However, this method does not work well for general nonlinear $f$. To see this, denote $z = f^{-1}(y) = f^{-1}[f(\mathbf{x}^\top \boldsymbol{\beta}^*) + \epsilon]$ and $\mu = \mathbb{E}[z | \mathbf{x}] - \mathbf{x}^\top \boldsymbol{\beta}^*$. Then we have model

$$z = \mathbf{x}^\top \boldsymbol{\beta}^* + \mu + \xi, \tag{2.5}$$

where $\xi$ is the remaining term that satisfies $\mathbb{E}[\xi | \mathbf{x}] = 0$. Note that both $\mu$ and $\xi$ depend on $\boldsymbol{\beta}^*$ implicitly. When treating (2.5) as a sparse linear model with intercept, we discard such dependency and thus incur large estimation error. We numerically compare the proposed method with the linear approach that inverts $f$ in §5 and show that our approach outperforms the linear framework.

## 2.2 Asymptotic Inference

Next, we consider asymptotic inference for the low-dimensional component of the true parameter $\boldsymbol{\beta}^*$. In particular, we consider the hypothesis testing problem $H_0: \beta_j^* = 0$ versus $H_1: \beta_j^* \neq 0$ for $j \in \{1, \ldots, d\}$. For some fixed $j$, we write $\boldsymbol{\beta}^* = (\alpha^*, \boldsymbol{\gamma}^{*\top})^\top$ with $\alpha^* = \beta_j^*$ and $\boldsymbol{\gamma}^* = (\beta_1^*, \ldots, \beta_{j-1}^*, \beta_{j+1}^*, \ldots, \beta_d^*)^\top \in \mathbb{R}^{d-1}$. We also define the partition $\boldsymbol{\beta} = (\alpha, \boldsymbol{\gamma}^\top)^\top$ in a similar fashion. Based on the gradient of $L(\boldsymbol{\beta})$, we construct the decorrelated score and Wald test statistics. Both of them are asymptotically normal distributed, yielding valid hypothesis tests and valid confidence intervals for $\alpha^*$. Here by saying the proposed hypothesis tests and confidence intervals are "valid", we mean the nominal type-I error and the coverage probability hold for the hypothesis tests and confidence intervals respectively when the sample size tends to infinity.

## 2.3 Decorrelated Score Test

For $\boldsymbol{\beta} = (\alpha, \boldsymbol{\gamma}^\top)^\top \in \mathbb{R}^d$ and $\rho > 0$, we define a decorrelated score function associated with $\alpha^*$ as

$$F_S(\boldsymbol{\beta}, \rho) := \nabla_\alpha L(\boldsymbol{\beta}) - \mathbf{d}(\boldsymbol{\beta}, \rho)^\top \nabla_{\boldsymbol{\gamma}} L(\boldsymbol{\beta}), \tag{2.6}$$

where $\mathbf{d}(\boldsymbol{\beta}, \rho) \in \mathbb{R}^{d-1}$ is constructed by the following Dantzig selector (Candès and Tao, 2007)

$$\mathbf{d}(\boldsymbol{\beta}, \rho) = \min_{\mathbb{R}^{d-1}} \|\mathbf{v}\|_1 \text{ subject to } \left\|\nabla^2_{\alpha\boldsymbol{\gamma}} L(\boldsymbol{\beta}) - \mathbf{v}^\top \nabla^2_{\boldsymbol{\gamma\gamma}} L(\boldsymbol{\beta})\right\|_\infty \le \rho, \tag{2.7}$$

and $L(\boldsymbol{\beta})$ is given in (2.1). Here $\rho$ is the second tuning parameter and $\nabla^2_{\alpha\boldsymbol{\gamma}} L(\boldsymbol{\beta})$ and $\nabla^2_{\boldsymbol{\gamma\gamma}} L(\boldsymbol{\beta})$ are the submatrices of $\nabla^2 L(\boldsymbol{\beta}) \in \mathbb{R}^{d \times d}$, which is the Hessian of $L(\boldsymbol{\beta})$. Here we denote

$$\nabla^2_{\alpha\alpha} L(\boldsymbol{\beta}) = \frac{\mathrm{d}^2 L(\boldsymbol{\beta})}{\mathrm{d}\alpha^2} \in \mathbb{R}, \quad \nabla^2_{\alpha\boldsymbol{\gamma}} L(\boldsymbol{\beta}) = \frac{\mathrm{d}^2 L(\boldsymbol{\beta})}{\mathrm{d}\alpha \, \mathrm{d}\boldsymbol{\gamma}} \in \mathbb{R}^{d-1}, \quad \nabla^2_{\boldsymbol{\gamma\gamma}} L(\boldsymbol{\beta}) = \frac{\mathrm{d}^2 L(\boldsymbol{\beta})}{\mathrm{d}\boldsymbol{\gamma} \, \mathrm{d}\boldsymbol{\gamma}} \in \mathbb{R}^{(d-1) \times (d-1)}.$$

We then define

$$\widehat{\sigma}_S^2(\boldsymbol{\beta}, \rho) := \left\{\frac{1}{n}\sum_{i=1}^n f'(\mathbf{x}_i^\top \boldsymbol{\beta})^2 (\mathbf{x}_i^\top \boldsymbol{w})^2\right\} \cdot \left\{\frac{1}{n}\sum_{i=1}^n [y_i - f(\mathbf{x}_i^\top \boldsymbol{\beta})]^2\right\}, \tag{2.8}$$

where $\boldsymbol{w} = \left[1, -\mathbf{d}(\boldsymbol{\beta}, \rho)^\top\right]^\top \in \mathbb{R}^d$ and $f'$ is the derivative of $f$. Denote $\widehat{\boldsymbol{\beta}} = (\widehat{\alpha}, \widehat{\boldsymbol{\gamma}}^\top)^\top$ to be the attained $\ell_1$-regularized least-square estimator and $\widetilde{\boldsymbol{\beta}} = (0, \widehat{\boldsymbol{\gamma}}^\top)^\top$. Our decorrelated score statistic is defined as $\sqrt{n} \cdot F_S(\widetilde{\boldsymbol{\beta}}, \rho) / \widehat{\sigma}_S(\widetilde{\boldsymbol{\beta}}, \rho)$, where we use $\widetilde{\boldsymbol{\beta}}$ instead of $\widehat{\boldsymbol{\beta}}$ since under the null hypothesis we have $\alpha^* = 0$. We will show in §3 that this quantity is asymptotically normal. Therefore the level-$\delta$ score test for testing $H_0: \alpha^* = 0$ versus $H_1: \alpha^* \ne 0$ is

$$\psi_S(\delta) := \begin{cases} 1 & \text{if } \left|\sqrt{n} \cdot F_S(\widetilde{\boldsymbol{\beta}}, \rho)/\widehat{\sigma}_S(\widetilde{\boldsymbol{\beta}}, \rho)\right| > \Phi^{-1}(1 - \delta/2) \\ 0 & \text{if } \left|\sqrt{n} \cdot F_S(\widetilde{\boldsymbol{\beta}}, \rho)/\widehat{\sigma}_S(\widetilde{\boldsymbol{\beta}}, \rho)\right| \le \Phi^{-1}(1 - \delta/2) \end{cases}, \tag{2.9}$$

where $\Phi^{-1}$ is the inverse cumulative distribution function of standard normal. The test rejects the null hypothesis if and only if $\psi_S(\delta) = 1$. In addition, the associated $p$-value takes the form

$$p\text{-value} = 2 \cdot \left[1 - \Phi\Big(\big|\sqrt{n} \cdot F_S(\widetilde{\boldsymbol{\beta}}, \rho)/\widehat{\sigma}_S(\widetilde{\boldsymbol{\beta}}, \rho)\big|\Big)\right].$$

To understand the intuition behind the above score test, we consider the case of Gaussian noise $\epsilon_i \sim N(0, \sigma^2)$, which implies that $y_i | \mathbf{x}_i \sim N\big(f(\mathbf{x}_i^\top \boldsymbol{\beta}^*), \sigma^2\big)$. Hence the negative log-likelihood function of $y_1, \ldots, y_n$ is exactly $L(\boldsymbol{\beta})/\sigma^2$ and the score function is proportional to $\nabla L(\boldsymbol{\beta})$. As discussed in Ning and Liu (2014a), in the presence of the high-dimensional nuisance parameter $\boldsymbol{\gamma}^*$, the classical score test is no longer valid. In detail, consider the following Taylor expansion of $\nabla_\alpha L$

$$\sqrt{n} \cdot \nabla_\alpha L(0, \bar{\boldsymbol{\gamma}}) = \sqrt{n} \cdot \nabla_\alpha L(0, \boldsymbol{\gamma}^*) + \sqrt{n} \cdot \nabla^2_{\alpha\boldsymbol{\gamma}} L(0, \boldsymbol{\gamma}^*)(\bar{\boldsymbol{\gamma}} - \boldsymbol{\gamma}^*) + \text{Rem}, \tag{2.10}$$

where $\bar{\boldsymbol{\gamma}}$ is an estimator of $\boldsymbol{\gamma}^*$ and Rem is the remainder term. In low dimensions, $\bar{\boldsymbol{\gamma}}$ is the maximum likelihood estimator and the asymptotic normality of $\sqrt{n} \cdot \nabla_\alpha L(0, \bar{\boldsymbol{\gamma}})$ in (2.10) can be implied by the fact that $\sqrt{n} \cdot \nabla_\alpha L(0, \boldsymbol{\gamma}^*)$ and $\sqrt{n} \cdot \nabla^2_{\alpha\boldsymbol{\gamma}} L(0, \boldsymbol{\gamma}^*)(\bar{\boldsymbol{\gamma}} - \boldsymbol{\gamma}^*)$ converge weakly to a jointly normal distribution and that Rem in (2.10) is $o_\mathbb{P}(1)$. In high dimensions, $\bar{\boldsymbol{\gamma}}$ is a sparse estimator, for example, the Lasso estimator, the limiting distribution of $\sqrt{n} \cdot \nabla^2_{\alpha\boldsymbol{\gamma}} L(0,, \boldsymbol{\gamma}^*)(\bar{\boldsymbol{\gamma}} - \boldsymbol{\gamma}^*)$ is no longer normal and becomes intractable (Knight and Fu, 2000).

We show in the next section that the decorrelated score defined in (2.6) overcomes the above difficulties. For any $\mathbf{v} \in \mathbb{R}^{d-1}$, by Taylor expansion, for any estimator $\bar{\boldsymbol{\gamma}}$ of $\boldsymbol{\gamma}^*$, we have

$$\sqrt{n} \cdot \nabla_\alpha L(0, \bar{\boldsymbol{\gamma}}) - \sqrt{n} \cdot \mathbf{v}^\top \nabla_{\boldsymbol{\gamma}} L(0, \bar{\boldsymbol{\gamma}}) = \overbrace{\sqrt{n} \cdot \nabla_\alpha L(0, \boldsymbol{\gamma}^*) - \sqrt{n} \cdot \mathbf{v}^\top \nabla_{\boldsymbol{\gamma}} L(0, \boldsymbol{\gamma}^*)}^{(i)}$$
$$+ \underbrace{\sqrt{n} \cdot \left[ \nabla^2_{\alpha\boldsymbol{\gamma}} L(0, \boldsymbol{\gamma}^*) - \mathbf{v}^\top \nabla^2_{\boldsymbol{\gamma\gamma}} L(0, \boldsymbol{\gamma}^*) \right] \cdot (\bar{\boldsymbol{\gamma}} - \boldsymbol{\gamma}^*)}_{(ii)} + \text{Rem}, \qquad (2.11)$$

where Rem is the remainder term. Note that if $\mathbf{v}$ is chosen such that $\mathbf{v}^\top \nabla^2_{\boldsymbol{\gamma\gamma}} L(0, \boldsymbol{\gamma}^*) = \nabla^2_{\alpha\boldsymbol{\gamma}} L(0, \boldsymbol{\gamma}^*)$, term (ii) in (2.11) becomes zero. Moreover, for term (i), the asymptotic normality is often guaranteed by the central limit theorem; the remainder term is $o_\mathbb{P}(1)$ if $\bar{\boldsymbol{\gamma}}$ has a fast rate of convergence. Hence, under these conditions we have that $\sqrt{n} \cdot \nabla_\alpha L(0, \bar{\boldsymbol{\gamma}}) - \sqrt{n} \cdot \mathbf{v}^\top \nabla_{\boldsymbol{\gamma}} L(0, \bar{\boldsymbol{\gamma}})$ is asymptotically normal. However, when $\boldsymbol{\gamma}^*$ is unknown, it is impossible to evaluate $\nabla^2 L(0, \boldsymbol{\gamma}^*)$. Instead we use $\nabla^2 L(\widetilde{\boldsymbol{\beta}})$ as an approximation. By the definition of the Dantzig selector in (2.7), term (ii) in (2.11) with $\mathbf{v}$ replaced by $\mathbf{d}(\widetilde{\boldsymbol{\beta}}, \rho)$ is negligible when $\rho$ is properly chosen. Since $\widehat{\boldsymbol{\gamma}}$ has a fast rate of convergence, we conclude that $\sqrt{n} \cdot F_S(\widetilde{\boldsymbol{\beta}}, \rho)$ is asymptotically normal since term (i) is a rescaled average of i.i.d. random variables for which the central limit theorem holds. Moreover, we also show that $\widehat{\sigma}^2_S(\widetilde{\boldsymbol{\beta}}, \rho)$ defined in (2.8) is a consistent estimator of the asymptotic variance of $\sqrt{n} \cdot F_S(\widetilde{\boldsymbol{\beta}}, \rho)$, which concludes the proof that the proposed score statistic in (2.6) is asymptotically standard normal.

It is worthy to note that Ning and Liu (2014b) propose a similar framework for generalized linear models. Compared to our results, their approach is likelihood-based and can only handle the case where the variance of the response variable $y$ is known. In addition, our inferential methods deal with a nonconvex loss function whereas in their case the loss function must be convex.

## 2.4 Construction of the Confidence Set

Given the decorrelated score statistic in (2.6), we further construct a confidence set for a low-dimensional component of $\boldsymbol{\beta}^*$. For $\boldsymbol{\beta} = (\alpha, \boldsymbol{\gamma}^\top)^\top \in \mathbb{R}^d$ and $\rho > 0$, we define

$$\bar{\alpha}(\boldsymbol{\beta}, \rho) = \widehat{\alpha} - \left\{ \nabla^2_{\alpha\alpha} L(\boldsymbol{\beta}) - \nabla^2_{\alpha\boldsymbol{\gamma}} L(\boldsymbol{\beta}) \cdot \mathbf{d}(\boldsymbol{\beta}, \rho) \right\}^{-1} \cdot F_S(\boldsymbol{\beta}, \rho), \qquad (2.12)$$

where $\mathbf{d}(\boldsymbol{\beta}, \rho)$ is obtained from the Dantzig selector (2.7) and $F_S(\boldsymbol{\beta}, \rho)$ is defined in (2.6). Let

$$\widehat{\sigma}^2_W(\boldsymbol{\beta}, \rho) := \left\{ \frac{1}{n} \sum_{i=1}^n f'(\mathbf{x}_i^\top \boldsymbol{\beta})^2 (\mathbf{x}_i^\top \boldsymbol{w})^2 \right\}^{-1} \cdot \left\{ \frac{1}{n} \sum_{i=1}^n [y_i - f(\mathbf{x}_i^\top \boldsymbol{\beta})]^2 \right\}, \qquad (2.13)$$

where $\boldsymbol{w} = \left[ 1, -\mathbf{d}(\boldsymbol{\beta}, \rho)^\top \right]^\top \in \mathbb{R}^d$. Let $\widehat{\boldsymbol{\beta}} = (\widehat{\alpha}, \widehat{\boldsymbol{\gamma}}^\top)^\top$ be the $\ell_1$-regularized least-square estimator obtained in (1.2). We define a Wald-type statistic as $\sqrt{n} \cdot \left[ \bar{\alpha}(\widehat{\boldsymbol{\beta}}, \rho) - \alpha^* \right] / \widehat{\sigma}_W(\widehat{\boldsymbol{\beta}}, \rho)$. We show in §3 that this quantity is asymptotically $N(0, 1)$. Then for any $\delta \in [0, 1]$, the asymptotic level-$(1 - \delta)$ confidence interval for $\alpha^*$ is given by

$$\left[ \bar{\alpha}(\widehat{\boldsymbol{\beta}}, \rho) - \Phi^{-1}(1 - \delta/2) \cdot \widehat{\sigma}_W(\widehat{\boldsymbol{\beta}}, \rho)/\sqrt{n}, \ \ \bar{\alpha}(\widehat{\boldsymbol{\beta}}, \rho) + \Phi^{-1}(1 - \delta/2) \cdot \widehat{\sigma}_W(\widehat{\boldsymbol{\beta}}, \rho)/\sqrt{n} \right]. \qquad (2.14)$$

The validity of the confidence interval can be intuitively understood as follows. Similar to the idea of the decorrelated score test, we can establish the asymptotic normality of $\sqrt{n} \cdot F_S(\widehat{\boldsymbol{\beta}}, \rho)$. Based on the classical Wald test (van der Vaart, 2000), we can further establish the asymptotic normality of $\sqrt{n} \cdot (\underline{\alpha} - \alpha^*)$, where $\underline{\alpha}$ is the solution to

$$S(\alpha, \widehat{\boldsymbol{\gamma}}; \widehat{\mathbf{d}}) := \nabla_\alpha L(\alpha, \widehat{\boldsymbol{\gamma}}) - \widehat{\mathbf{d}}^\top \nabla_{\boldsymbol{\gamma}} L(\alpha, \widehat{\boldsymbol{\gamma}}) = 0, \ \ \text{where } \widehat{\mathbf{d}} = \mathbf{d}(\widehat{\boldsymbol{\beta}}, \rho). \qquad (2.15)$$

However, (2.15) may have multiple roots, which makes the estimator ill-posed. Instead of solving for $\underline{\alpha}$ directly, we consider the first-order approximation of (2.15):

$$S(\alpha, \widehat{\boldsymbol{\gamma}}; \widehat{\mathbf{d}}) + (\alpha - \widehat{\alpha}) \cdot \nabla_\alpha S(\alpha, \widehat{\boldsymbol{\gamma}}; \widehat{\mathbf{d}}) = 0, \tag{2.16}$$

where $\nabla_\alpha S(\alpha, \widehat{\boldsymbol{\gamma}}; \widehat{\mathbf{d}}) = \nabla_{\alpha\alpha} L(\widehat{\boldsymbol{\beta}}) - \widehat{\mathbf{d}}^\top \nabla_{\boldsymbol{\gamma}\alpha} L(\widehat{\boldsymbol{\beta}})$ and $\widehat{\boldsymbol{\beta}} = (\widehat{\alpha}, \widehat{\boldsymbol{\gamma}}^\top)^\top$. Hence, $\bar{\alpha}(\boldsymbol{\beta}, \rho)$ defined in (2.12) is the unique solution of (2.16) and we can intuitively view $\bar{\alpha}(\boldsymbol{\beta}, \rho)$ as an approximation of $\underline{\alpha}$. Therefore, we expect $\sqrt{n} \cdot [\bar{\alpha}(\boldsymbol{\beta}, \rho) - \alpha^*]$ to be asymptotically normal. In addition, we also show that $\widehat{\sigma}_W^2(\widehat{\boldsymbol{\beta}}, \rho)$ is a consistent estimator of the asymptotic variance of $\sqrt{n} \cdot [\bar{\alpha}(\boldsymbol{\beta}, \rho) - \alpha^*]$, which guarantees the validity of the confidence interval in (2.14).

As a byproduct, we show that the Wald-type statistic $\sqrt{n} \cdot [\bar{\alpha}(\widehat{\boldsymbol{\beta}}, \rho) - \alpha^*]/\widehat{\sigma}_W(\widehat{\boldsymbol{\beta}}, \rho)$ also yields a valid hypothesis test. By the asymptotic normality of this statistic, the level-$\delta$ high-dimensional Wald test for testing $H_0: \alpha^* = 0$ versus $H_1: \alpha^* \neq 0$ is defined as

$$\psi_W(\delta) := \begin{cases} 1 & \text{if } \left|\sqrt{n} \cdot \bar{\alpha}(\widehat{\boldsymbol{\beta}}, \rho)/\widehat{\sigma}_W(\widehat{\boldsymbol{\beta}}, \rho)\right| > \Phi^{-1}(1 - \delta/2) \\ 0 & \text{if } \left|\sqrt{n} \cdot \bar{\alpha}(\widehat{\boldsymbol{\beta}}, \rho)/\widehat{\sigma}_W(\widehat{\boldsymbol{\beta}}, \rho)\right| \leq \Phi^{-1}(1 - \delta/2), \end{cases} \tag{2.17}$$

where $\Phi^{-1}$ is the inverse function of the cumulative distribution function of the standard normal distribution. The test rejects the null hypothesis if and only if $\psi_W(\delta) = 1$. In addition, the associated $p$-value takes on the form

$$p\text{-value} = 2 \cdot \left[1 - \Phi\left(\left|\sqrt{n} \cdot \bar{\alpha}(\widehat{\boldsymbol{\beta}}, \rho)/\widehat{\sigma}_W(\widehat{\boldsymbol{\beta}}, \rho)\right|\right)\right].$$

## 3 Theoretical Results

In this section, we present the main theoretical results. The statistical model is defined in (1.1). Hereafter we assume that $\epsilon$ is sub-Gaussian with variance proxy $\sigma^2$. By saying that a random vector $\mathbf{z} \in \mathbb{R}^k$ is sub-Gaussian with zero mean and variance proxy $\tau^2 \geq 0$, we mean that

$$\mathbb{E}[\mathbf{z}] = \mathbf{0} \quad \text{and} \quad \mathbb{E}[\exp(\boldsymbol{\theta}^\top \mathbf{z})] \leq \exp(\|\boldsymbol{\theta}\|_2^2 \tau^2 / 2) \quad \text{for all } \boldsymbol{\theta} \in \mathbb{R}^k.$$

### 3.1 Theory of Parameter Estimation

Before presenting the main results for parameter estimation, we first state the following assumptions on $\widehat{\boldsymbol{\Sigma}} = n^{-1} \sum_{i=1}^n \mathbf{x}_i \mathbf{x}_i^\top$, which are standard for sparse linear regression problems with fixed design.

**Assumption 1.** Sparse-Eigenvalue($s^*, k^*$). For $k \in \{1, \ldots, d\}$, we denote the $k$-sparse eigenvalues of $\widehat{\boldsymbol{\Sigma}}$ as $\rho_-(k)$ and $\rho_+(k)$ respectively, which are defined as

$$\rho_-(k) := \inf\{\mathbf{v}^\top \widehat{\boldsymbol{\Sigma}} \mathbf{v} : \|\mathbf{v}\|_0 \leq k, \|\mathbf{v}\|_2 = 1\} \quad \text{and} \quad \rho_+(k) := \sup\{\mathbf{v}^\top \widehat{\boldsymbol{\Sigma}} \mathbf{v} : \|\mathbf{v}\|_0 \leq k, \|\mathbf{v}\|_2 = 1\}.$$

We assume that, for $s^* = \|\boldsymbol{\beta}^*\|_0$, there exists a $k^* \in \mathbb{N}$ such that

$$\rho_+(k^*)/\rho_-(2k^* + s^*) \leq 1 + 0.5k^*/s^* \quad \text{and} \quad k^* \geq 2s^*. \tag{3.1}$$

The condition $\rho_+(k^*)/\rho_-(2k^* + s^*) \leq 1 + 0.5k^*/s^*$ requires that the eigenvalue ratio $\rho_+(k)/\rho_-(2k + s^*)$ grows sub-linearly in $k$. This condition, commonly referred to as *sparse eigenvalue condition*, is standard in sparse estimation problems and has been studied by Zhang (2010). This condition is weaker than the well-known restricted isometry property (RIP) in compressed sensing (Candès and Tao, 2005), which states that there exists a constant $\delta \in (0, 1)$ and integer $s \in \{1, \ldots, d\}$ such that

$$(1 - \delta)\|\mathbf{v}\|_2^2 \leq \mathbf{v}^\top \widehat{\boldsymbol{\Sigma}} \boldsymbol{v} \leq (1 + \delta)\|\mathbf{v}\|_2^2 \text{ for all } \mathbf{v} \in \mathbb{R}^d \text{ satisfying } \|\mathbf{v}\|_0 \leq s. \tag{3.2}$$

Comparing (3.1) and (3.2), we see that (3.1) holds with $k^* = (s - s^*)/2$ if the RIP condition holds with $s \geq 5s^*$ and $\delta = 1/3$. As is shown in Vershynin (2010), RIP holds with high probability for sub-Gaussian random matrices. Therefore Assumption 1 holds at least when $\mathbf{x}_1, \ldots, \mathbf{x}_n$ are i.i.d. sub-Gaussian, which contains many well-known distributions as special cases.

We note that although Assumption Sparse-Eigenvalue$(s^*, k^*)$ holds since it does not depend on the nonlinear transfmration $f$, the *restricted strong convexity* (RSC) condition defined in Loh and Wainwright (2014, 2015) on the loss function $L(\boldsymbol{\beta})$ does not directly hold in general in our setting since $L(\boldsymbol{\beta})$ depends on the nonlinear transformation $f$.

In addition to the sparse eigenvalue assumption, we need a regularity condition, which states that the elements of $\widehat{\boldsymbol{\Sigma}}$ are uniformly bounded.

**Assumption 2.** Bounded-Design$(D)$. We assume there exists an absolute constant $D$ that does not depends on $n, d$, or $s^*$ such that $\|\widehat{\boldsymbol{\Sigma}}\|_\infty \leq D$, where $\|\cdot\|_\infty$ is the matrix elementwise $\ell_\infty$-norm.

If the population version of $\widehat{\boldsymbol{\Sigma}}$, i.e., $\boldsymbol{\Sigma} := \mathbb{E}(\mathbf{x}\mathbf{x}^\top)$, has bounded elements and $\mathbf{x}$ has sub-Gaussian or sub-exponential tails, then by concentration inequalities we can prove that Assumption Bounded-Design$(D)$ holds with high probability with $D = 2\|\boldsymbol{\Sigma}\|_\infty$. We will verify this assumption for sub-Gaussian $\mathbf{x}$ in the appendix. This assumption is generally unnecessary for high dimensional linear regression. However, it is required in our setting where it is used to control the effect of the nonlinear transform.

We note that we do not make any further assumptions except Assumptions 1 and 2 on the distribution of $\mathbf{x}$ for the theory of parameter estimation to hold. These two assumptions are shown to be true when $\mathbf{x}$ is sub-Gaussian.

We are now ready to present our main theorem for parameter estimation, which states that any stationary point of the $\ell_1$-regularized optimization problem enjoys optimal statistical rates of convergence and that Algorithm 1 successfully converges to a stationary point.

**Theorem 1.** We assume that the univariate function $f$ satisfies $f(0) = 0$ and is continuously differentiable with $f'(x) \in [a, b], \forall x \in \mathbb{R}$ for some $0 < a < b$. We further assume that Assumptions 1 and 2 hold. Then there exists a constant $B$ such that $\|\nabla L(\boldsymbol{\beta}^*)\|_\infty \leq B\sigma \cdot \sqrt{\log d / n}$ with probability tending to one. Suppose we choose the regularization parameter $\lambda$ in (1.2) as

$$\lambda = C\sigma\sqrt{\log d/n} \quad \text{with} \quad C \geq \max\{L_1 B, L_2\}, \quad (3.3)$$

where $L_1$ and $L_2$ satisfy $L_1^{-1} + 3b\sqrt{D}L_2^{-1} \leq 0.1$. Then for any stationary point $\widehat{\boldsymbol{\beta}}$ satisfying $\nabla L(\widehat{\boldsymbol{\beta}}) + \lambda \cdot \boldsymbol{\xi} = \mathbf{0}$ with $\boldsymbol{\xi} \in \partial \|\widehat{\boldsymbol{\beta}}\|_1$, it holds with probability at least $1 - d^{-1}$ that

$$\|\widehat{\boldsymbol{\beta}} - \boldsymbol{\beta}^*\|_2 \leq 25/\rho_-(k^* + s^*) \cdot a^{-2}\sqrt{s^*}\lambda \text{ and } \|\widehat{\boldsymbol{\beta}} - \boldsymbol{\beta}^*\|_1 \leq 25/\rho_-(k^* + s^*) \cdot a^{-2}s^*\lambda. \quad (3.4)$$

Furthermore, Algorithm 1 attains a stationary point with the statistical rates in (3.4).

By our discussion under Assumption 1, we can take $k^* = Cs^*$ for some constant $C > 0$. Then plugging (3.3) into (3.4), we obtain the rate of $\sqrt{s^* \log d/n}$ in $\ell_2$-norm and the rate of $s^*\sqrt{\log d/n}$ in $\ell_1$-norm. Similar results are also established for sparse linear regression, and more generally, high-dimensional generalized linear models (Candès and Tao, 2007; Zhang and Huang, 2008; Kakade et al., 2010). These rates are optimal in the sense that they cannot be improved even if $f$ equals to the identity. Note that the lower bound $a$ of $f'$ shows up in the statistical rates of convergence in (3.4). If $a$ is close to zero, we obtain a large statistical error. To see the intuition, we consider a worst case where $f$ is constant, i.e., $a = 0$. Then it is impossible to consistently estimate $\boldsymbol{\beta}^*$, since in this case the observations $\{y_i, \mathbf{x}_i\}_{i=1}^n$ provide no information on $\boldsymbol{\beta}^*$.

The statistical rates of convergence are proportional to the noise level $\sigma$, which implies that the proposed method exactly recovers $\boldsymbol{\beta}^*$ in the noiseless setting. In the noisy case, by (3.4), to obtain $\epsilon$ accuracy of estimating $\boldsymbol{\beta}^*$ in $\ell_2$-norm with high probability, the sample complexity is $n = O(\epsilon^{-2} s^* \log d)$, which is of the same order as that of high-dimensional linear models.

To understand the optimality of the estimation result, we study the minimax lower bound of estimation in our model, which reveals the fundamental limits of the estimation problem. We define the minimax risk as

$$\mathcal{R}_f^*(s, n, d) = \inf_{\widehat{\boldsymbol{\beta}}} \sup_{\boldsymbol{\beta} \in \mathbb{B}_0(s)} \mathbb{E}_{\boldsymbol{\beta}} \|\widehat{\boldsymbol{\beta}} - \boldsymbol{\beta}\|_2^2, \tag{3.5}$$

where the expectation is taken over the probability model in (1.1) with parameter $\boldsymbol{\beta}$ and $\mathbb{B}_0(s) := \{\boldsymbol{\beta} \in \mathbb{R}^d \colon \|\boldsymbol{\beta}\|_0 \leq s\}$. Here the supremum is taken over all $s$-sparse parameters and the infimum is taken over all estimators $\widehat{\boldsymbol{\beta}}$ based on samples $\{(y_i, \mathbf{x}_i)\}_{i=1}^n$. We assume $f$ is continuously differentiable with $f'(u) \in [a, b], \forall u \in \mathbb{R}$. The following theorem gives a lower bound on the minimax risk $\mathcal{R}_f(s, n, d)$, which implies the optimality of the proposed estimator.

**Theorem 2.** For integer $s$ and $d$ satisfying $1 \leq s \leq d/8$, the minimax risk defined in (3.5) has the following lower bound

$$\mathcal{R}_f^*(s, n, d) \geq \frac{\sigma^2}{192 b^2 \rho_+(2s)} \frac{s \log[1 + d/(2s)]}{n}. \tag{3.6}$$

By Theorem 2, if we consider $a, b$ as constants and assume that $k^*/s^*$ is bounded, then the $\ell_2$-statistical rate of convergence of $\widehat{\boldsymbol{\beta}}$ in (3.4) matches the minimax lower bound in (3.6) in terms of order. This establishes the optimality of the proposed estimator.

## 3.2 Theory of Asymptotic Inference

As described in §2.2, we establish the theoretical results of asymptotic inference for the low-dimensional component $\alpha^*$ of $\boldsymbol{\beta}^*$, where we write $\boldsymbol{\beta}^* = (\alpha^*, \boldsymbol{\gamma}^*)$ with $\alpha^* \in \mathbb{R}$ and $\boldsymbol{\gamma}^* \in \mathbb{R}^{d-1}$. To have a non-degenerate Fisher information, we need stronger assumptions. We make the following assumption on $f$ and the distribution of $\mathbf{x}$.

**Assumption 3.** We assume that besides the assumptions on $f$ in Theorem 1, it holds that,

(i) $f$ is twice differentiable and that there exist a constant $R$ such that $|f''(x)| \leq R, \forall x \in \mathbb{R}$.

(ii) $f''$ is $L_f$-Lipschitz, that is, for any $x, y \in \mathbb{R}$, $|f''(x) - f''(y)| \leq L_f \cdot |x - y|$.

In addition, we assume $\mathbf{x} \in \mathbb{R}^d$ is a sub-Gaussian random vector with mean zero variance proxy $\sigma_x^2$, and $\|\mathbb{E}(\mathbf{x}\mathbf{x}^\top)\|_\infty$ is bounded by a constant.

Note that the proposed procedures are also applicable when the tail of $\mathbf{x}$ is heavier than sub-Gaussian. In this case, for asymptotic inference, however, the scaling between $n$ and $d$ will be different. We assume that $\mathbf{x}$ is sub-Gaussian here for the simplicity of presentation. In the appendix, we prove that Assumption 3 implies that Assumptions 1 and 2 hold with high probability.

In what follows, we present an assumption on some population quantities and sample size $n$. We denote $\mathbf{I}(\boldsymbol{\beta}^*) := \mathbb{E}[\nabla^2 L(\boldsymbol{\beta}^*)] = \mathbb{E}[f'(\mathbf{x}^\top \boldsymbol{\beta}^*)^2 \mathbf{x}\mathbf{x}^\top] \in \mathbb{R}^{d \times d}$, which is the Fisher information matrix. In our notations, $[\mathbf{I}(\boldsymbol{\beta}^*)]_{\boldsymbol{\gamma},\boldsymbol{\gamma}} \in \mathbb{R}^{(d-1) \times (d-1)}$ and $[\mathbf{I}(\boldsymbol{\beta}^*)]_{\boldsymbol{\gamma},\alpha} \in \mathbb{R}^{(d-1)}$ are the submatrices of $\mathbf{I}(\boldsymbol{\beta}^*)$. When $[\mathbf{I}(\boldsymbol{\beta}^*)]_{\boldsymbol{\gamma},\boldsymbol{\gamma}}$ is invertible, let

$$\mathbf{d}^* := [\mathbf{I}(\boldsymbol{\beta}^*)]_{\boldsymbol{\gamma},\boldsymbol{\gamma}}^{-1} [\mathbf{I}(\boldsymbol{\beta}^*)]_{\boldsymbol{\gamma},\alpha}, \quad s_{\mathbf{d}}^* := \|\mathbf{d}^*\|_0, \quad \text{and} \quad \mathcal{S}_{\mathbf{d}} := \text{supp}(\mathbf{d}^*). \tag{3.7}$$

In addition, we define $\sigma_S^2(\boldsymbol{\beta}^*) := \mathbb{E}(\epsilon^2) \sigma^2(\boldsymbol{\beta}^*)$ where $\sigma^2(\boldsymbol{\beta}^*) = [1, -(\mathbf{d}^*)^\top] \mathbf{I}(\boldsymbol{\beta}^*) [1, -(\mathbf{d}^*)^\top]^\top$. The next assumption ensures the eigenvalues of $[\mathbf{I}(\boldsymbol{\beta}^*)]_{\boldsymbol{\gamma},\boldsymbol{\gamma}}$ and $\sigma(\boldsymbol{\beta}^*)$ are asymptotically bounded from above and below.

**Assumption 4.** We assume that there exist positive absolute constants $\tau^*$ and $\tau_*$ such that

$$\tau^* \geq \lambda_1[\mathbf{I}(\boldsymbol{\beta}^*)] \geq \lambda_d[\mathbf{I}(\boldsymbol{\beta}^*)] \geq \tau_* > 0.$$

We also assume that $\sigma^2(\boldsymbol{\beta}^*) = \mathcal{O}(1)$ and $1/\sigma^2(\boldsymbol{\beta}^*) = \mathcal{O}(1)$. Furthermore, the tuning parameter $\rho$ of the Dantzig selector in (2.7) is set to $C(1 + \|\mathbf{d}^*\|_1)s^*\lambda$, where $\lambda$ appears in (1.2) and $C \geq 1$ is a sufficiently large absolute constant. We further assume that the sample size $n$ is assumed to be sufficiently large such that

$$\log^5 d/n = \mathcal{O}(1), \quad (1 + \|\mathbf{d}^*\|_1) \cdot s_{\mathbf{d}}^* \cdot \rho = o(1) \quad \text{and} \quad (s^* + s_{\mathbf{d}}^*) \cdot \lambda \cdot \rho = o(1/\sqrt{n}).$$

Recall that in Theorem 1 we set $\lambda \asymp \sqrt{\log d/n}$. Therefore Assumption 4 requires

$$\log^5 d/n = \mathcal{O}(1), \quad (1 + \|\mathbf{d}^*\|_1)^2 s^* s_{\mathbf{d}}^* \sqrt{\log d/n} = o(1) \quad \text{and} \quad (s^* + s_{\mathbf{d}}^*)(1 + \|\mathbf{d}^*\|_1)s^* \log d/\sqrt{n} = o(1).$$

If we treat $\|\mathbf{d}^*\|_1$ as constant and assume that $s_{\mathbf{d}}^*/s^* = \mathcal{O}(1)$, then Assumption 4 essentially implies

$$\log^5 d/n = \mathcal{O}(1) \quad \text{and} \quad s^{*4} \log^2 d/n = o(1).$$

**Theorem 3.** Under Assumptions 3 and 4, it holds that for $n \to \infty$,

$$\sqrt{n} \cdot F_S(\widehat{\boldsymbol{\beta}}', \rho)/\widehat{\sigma}_S(\widehat{\boldsymbol{\beta}}', \rho) \rightsquigarrow N(0,1),$$

where $\widehat{\boldsymbol{\beta}}' = (0, \widehat{\boldsymbol{\gamma}}^\top)^\top$ is obtained by replacing $\widehat{\alpha}$ with zero in a stationary point $\widehat{\boldsymbol{\beta}}$ of the $\ell_1$-regularized least-square optimization (1.2) and $\widehat{\sigma}_S(\widehat{\boldsymbol{\beta}}', \rho)$ is defined in (2.8). Furthermore, the level-$\delta$ score test defined in (2.9) achieves its nominal type-I error.

**Theorem 4.** We write $\boldsymbol{\beta}^* = (\alpha^*, \boldsymbol{\gamma}^\top)^\top$ where $\alpha^* \in \mathbb{R}$ is the low-dimensional parameter of interest. Under Assumptions 3 and 4, it holds that for $n \to \infty$,

$$\sqrt{n} \cdot [\bar{\alpha}(\widehat{\boldsymbol{\beta}}, \rho) - \alpha^*]/\widehat{\sigma}_W(\widehat{\boldsymbol{\beta}}, \rho) \rightsquigarrow N(0,1),$$

where $\widehat{\boldsymbol{\beta}} = (\widehat{\alpha}, \widehat{\boldsymbol{\gamma}}^\top)^\top$ a stationary point $\widehat{\boldsymbol{\beta}}$ of the $\ell_1$-regularized least-square optimization (1.2) and $\widehat{\sigma}_W(\widehat{\boldsymbol{\beta}}, \rho)$ is defined in (2.8). Therefore, the level-$(1-\delta)$ confidence interval defined in (2.14) has asymptotically $1-\delta$ covering probability. Moreover, the level-$\delta$ Wald test defined in (2.17) achieves its nominal type-I error.

## 4 Connection to Prior Work

The model we consider is closely related to the single index model where the function $f$ is unknown. Both of these two models fall in the framework of sufficient dimension reduction with a one-dimensional subspace $\mathcal{U}$ (Li, 1991, 1992; Cook, 1998; Cook and Lee, 1999; Li, 2007). In low dimensional settings, most works in this direction use spectral methods, which rely on the Gaussian assumption and can only estimate $\boldsymbol{\theta}^* = \boldsymbol{\beta}^*\|\boldsymbol{\beta}^*\|_2^{-1}$ because the norm of $\boldsymbol{\beta}^*$ is not identifiable when $f$ is unknown. As introduced in Li (2007), many moment based sufficient dimension reduction methods can be stated as a generalized eigenvalue problem $\mathbf{M}_n\boldsymbol{\theta}_i = \lambda_i\mathbf{N}_n\boldsymbol{\theta}_i$ for $i = 1, \ldots, d$, where $\mathbf{M}_n$ and $\mathbf{N}_n$ are symmetric matrices computed from the data; $\boldsymbol{\theta}_1, \ldots, \boldsymbol{\theta}_d$ are generalized eigenvectors such that $\boldsymbol{\theta}_i^\top \mathbf{N}_n \boldsymbol{\theta}_j = \mathbb{1}_{\{i=j\}}$ and $\lambda_1 \geq \cdots \geq \lambda_d$ are the generalized eigenvalues. In addition, it is required that $\mathbf{M}_n$ and $\mathbf{N}_n$ are positive semidefinite and positive definite, respectively. Here $\mathbf{M}_n$ and $\mathbf{N}_n$ are the sample versions of the corresponding population quantities $\mathbf{M}$ and $\mathbf{N}$. For example, in sliced inverse regression (Li, 1991), we have $\mathbf{M} = \text{Cov}\{\mathbb{E}[\mathbf{x} - \mathbb{E}(\mathbf{x})|y]\}$ and $\mathbf{N} = \text{Cov}(\mathbf{x})$ and $\mathbf{M}_n$ and $\mathbf{N}_n$ are their population analogs. When $\mathcal{U}$ is one-dimensional, $\boldsymbol{\theta}^*$ corresponds to the generalized eigenvector with the largest eigenvalue. In low dimensional settings, Li (2007) showed that $\boldsymbol{\theta}^*$ can be estimated by the following optimization problem:

$$\underset{\boldsymbol{\theta} \in \mathbb{R}^d}{\text{maximize}} \ \boldsymbol{\theta}^\top \mathbf{M}_n \boldsymbol{\theta} \quad \text{subject to } \boldsymbol{\theta}^\top \mathbf{N}_n \boldsymbol{\theta}. \tag{4.1}$$

Since the works in this direction all require the matrix $\mathbf{N}_n$, which is the sample covariance matrix of $\mathbf{x}$ in most cases, to be invertible, such methods cannot be generalized to high-dimensional settings where $\mathbf{N}_n$ is not invertible.

For high-dimensional single index models, Plan et al. (2014) proposes an estimator by projecting $n^{-1} \sum_{i=1}^n y_i \mathbf{x}_i$ onto a fixed star-shaped closed subset $K$ of $\mathbb{R}^d$. Similarly, Plan and Vershynin (2015) propose a least-squares estimator with a geometric constraint:

$$\text{minimize} \sum_{i=1}^n (\mathbf{x}_i^\top \boldsymbol{\theta} - y_i)^2 \ \text{ subject to } \ \boldsymbol{\theta} \in K. \tag{4.2}$$

Both of these methods rely on the assumption that $\mathbf{x}_i$ is Gaussian to have good estimation of $\mathbb{E}(y \cdot \mathbf{x})$. Under the Gaussian assumption, we achieve the same statistical rate, which is optimal. When $\mathbf{x}$ is not Gaussian, as shown in Ai et al. (2014), their methods will have some extra terms in the error bound that may or may not tend to zero. Our method, however, works when $\mathbf{x}$ has a general distribution with optimal statistical rates of convergence.

## 5 Numerical Experiments

In this section, we evaluate the finite sample performance of parameter estimation and asymptotic inference on simulated data.

For parameter estimation, we compute the $\ell_2$-error $\|\widehat{\boldsymbol{\beta}} - \boldsymbol{\beta}^*\|_2$, where $\widehat{\boldsymbol{\beta}}$ is the solution of Algorithm 1. In addition, we compare our method with the linear approach that inverts the nonlinear function. For the linear method we apply the $\ell_1$-regularized regresion (Lasso) (Tibshirani, 1996). For asymptotic inference, we report the type-1 error and power of the proposed method. Our numerical results corroborate with the previously established theories.

Throughout this section, we sample independent data from model (1.1) with $\epsilon \sim N(0,1)$ and $\mathbf{x} \sim N(\mathbf{0}, \boldsymbol{\Sigma})$ where $\boldsymbol{\Sigma} \in \mathbb{R}^{d \times d}$ is a Toeplitz matrix with $\Sigma_{jk} = 0.95^{|j-k|}$. The sparse parameter vector $\boldsymbol{\beta}^* \in \mathbb{R}^d$ is set to have nonzero values in the first $s^*$ entries. That is, $\beta_j^* \neq 0$ for $1 \leq j \leq s^*$ and $\beta_j^* = 0$ otherwise. In addition, we consider the nonlinear function $f(x) = 2x + \cos(x)$. In this case the derivative $f'(\cdot)$ is bounded by $a = 1$ and $b = 4$.

### 5.1 Parameter Estimation

For parameter estimation, we compare the $\ell_2$-error $\|\widehat{\boldsymbol{\beta}} - \boldsymbol{\beta}^*\|_2$ with $\sqrt{s^* \log d/n}$ under two settings: (i) we fix $d = 256$, $s^* = 6, 8$, or $10$, and vary $n$, and (2) fix $s^* = 10$, $d = 128, 256$ or $512$, and vary $n$. For the parameter $\boldsymbol{\beta}^*$, the first $s^*$ entries are sampled independently from the uniform distribution on the interval $[0, 2]$. That is, $\beta_j^* \sim \text{U}(0, 2)$ for $1 \leq j \leq s^*$ and $\beta_j^* = 0$ for $j > s^*$. We set the regularization parameter $\lambda = 3\sigma \cdot \sqrt{\log d/n}$. The parameters of Algorithm 1 are chosen as $\alpha_{\min} = 1/\alpha_{\max} = 10^{30}$, $\eta = 2$, $M = 5$, and $\zeta = \texttt{tol} = 10^{-5}$. The $\ell_2$-errors reported are based on 100 independent experiments. We plot the $\ell_2$-errors against the effective sample size $\sqrt{s^* \log d/n}$ in Figure 1. The figure illustrates that $\|\widehat{\boldsymbol{\beta}} - \boldsymbol{\beta}^*\|_2$ grows sublinearly with $\sqrt{s^* \log d/n}$, which corroborates with our argument that $\|\widehat{\boldsymbol{\beta}} - \boldsymbol{\beta}^*\|_2 \leq C\sqrt{s^* \log d/n}$ for some absolute constant $C$.

To compare Algorithm 1 with inverting $f$, we consider the settings where $d = 256$, $s^* = 8$. We then apply Lasso to the inverted data $\{f^{-1}(y_i), \mathbf{x}_i\}_{i=1}^n$ where the regularization parameter of Lasso is selected via 5-fold cross-validation. The optimization problem of Lasso is also solved using Algorithm 1. We plot the $\ell_1$-errors of these two techniques against the effective sample size in Figure 2, which shows that the proposed method outperforms the linear approach.

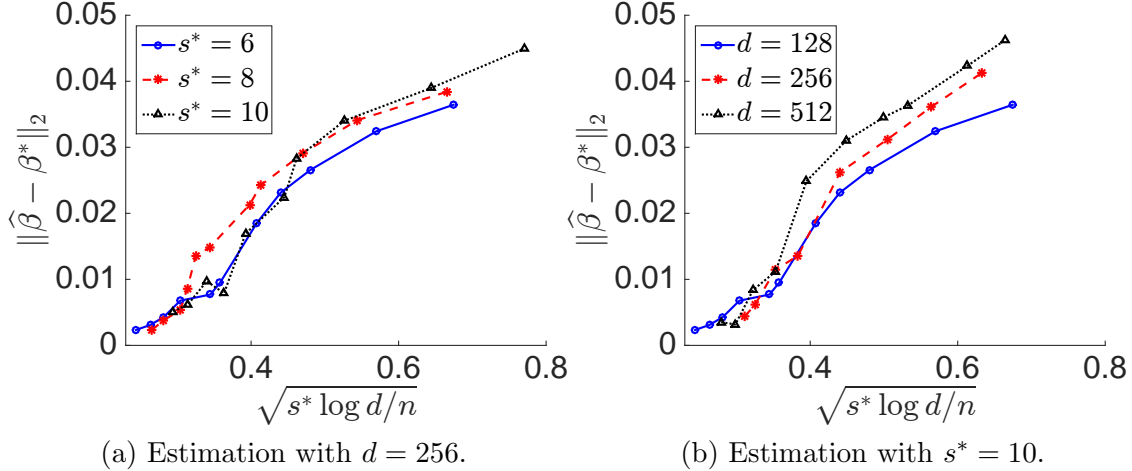

(a) Estimation with $d = 256$.  (b) Estimation with $s^* = 10$.

Figure 1: Statistical error $\|\widehat{\boldsymbol{\beta}} - \boldsymbol{\beta}^*\|_2$ plotted against the effective sample size $\sqrt{s^* \log d/n}$ with $d$ or $s^*$ fixed and $n$ varied.

## 5.2 Asymptotic Inference

To examine the finite sample performance of the proposed inferential procedures, we consider the setting where $n = 200$, $d = 512$ and $s^* = 10$. For $1 \leq j \leq s^*$, we let $\beta_j^* = \mu$ where $\mu \in \{0, 0.05, 0.1, \ldots, 1\}$. For $j > s^*$, we let $\beta_j^* = 0$. We consider the hypothesis testing problem $H_0\colon \beta_{11}^* = 0$ versus $H_0\colon \beta_{11}^* \neq 0$ and construct the decorrelated score and Wald tests with significance level $\delta = 0.05$. Note that here the null hypothesis is true. After repeated experiments, we regard the frequency of rejecting $H_0$ as the type-1 error of the proposed hypothesis tests. Moreover, we also consider testing $H_0\colon \beta_1^* = 0$ against $H_1\colon \beta_1^* \neq 0$. Since in this case the null hypothesis is not true, we use one minus the frequency of rejecting the null hypothesis as the power of these tests. The tuning parameter in the $\ell_1$-regularization problem (1.2) is set to be $\lambda = 3\sigma \cdot \sqrt{\log d/n}$ and the tuning parameter of the Dantzig selector (2.7) is set to be $\rho = 30\sigma \cdot \sqrt{\log d/n}$. We report the type-I errors and the powers of the proposed tests based on 500 independent trials. The results are listed in Table 1. As shown in the rows of the table, both the decorrelated score and Wald test controls the type-I error close to the significance level, which shows the validity of these tests empirically. In addition, we plot the powers of the proposed hypothesis tests against the value of $\mu$. As shown in Figure 3 and rows of Table 1, the powers of both tests increases as $\mu$ grows. This is not surprising because it would be easier to discriminate $\beta_1^*$ from zero if $\mu$ is large. Otherwise if $\mu$ is close to zero, it would be extremely difficult to determine whether $H_0\colon \beta_1^* = 0$ is true. In conclusion, the proposed decorrelated score and Wald test are valid for testing the low dimensional component of the high-dimensional signal $\boldsymbol{\beta}^*$ for our nonlinear regression problem.

## 6 Proof of the Main Results

In this section, we lay out the proofs of the main results presented in Section 3. We first establish the statistical rate of convergence for the proposed estimator, and then show the validity of our procedure for statistical inference. We leave the proofs of the auxiliary lemmas in the appendix.

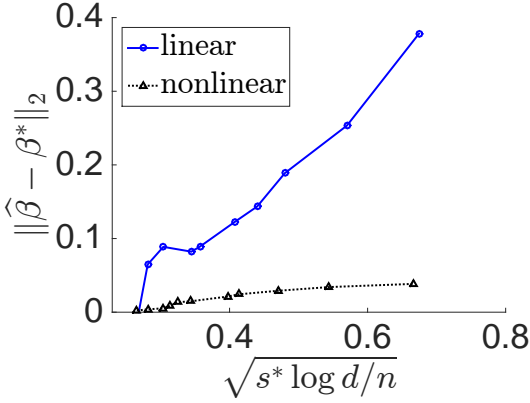
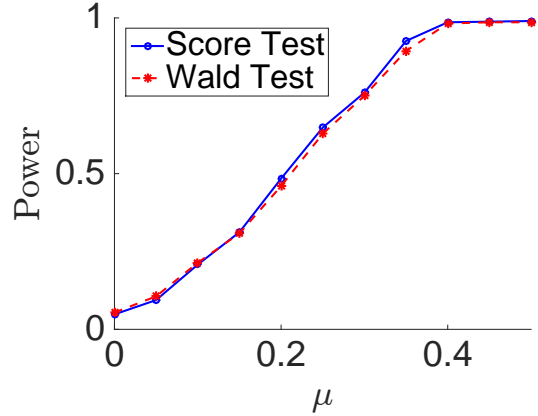

Figure 2: Statistical error $\|\widehat{\boldsymbol{\beta}} - \boldsymbol{\beta}^*\|_2$ plotted against the effective sample size $\sqrt{s^* \log d/n}$ for the proposed method and the linear method

Figure 3: Power curves of the proposed methods for testing $H_0\colon \beta_1^* = 0$ for the nonlinear sensing model at the 0.05 significance level.

### 6.1 Proof of Theorem 1

Before we introduce the proof of Theorem 1, we first introduce a technical condition which simplifies our presentation. This condition states that $\nabla L(\boldsymbol{\beta}^*)$ is of the order $\sigma \cdot \sqrt{\log d/n}$ with high probability.

**Condition 5** . `Bounded-Gradient`$(B, \delta)$. Let $L(\boldsymbol{\beta})$ be the least-square loss function defined in (2.1), there exist an absoulute constant $B > 0$ that does not depend on $n$, $d$ or $s^*$ and $\delta = \delta(n, d)$ that tends to 0 as $n \to \infty$ such that $\|\nabla L(\boldsymbol{\beta}^*)\|_\infty \leq B\sigma \cdot \sqrt{\log d/n}$ with probability at least $1 - \delta$.

To see why this condition holds, note that by definition we have,

$$\nabla L(\boldsymbol{\beta}^*) = -\frac{1}{n}\sum_{i=1}^n [y_i - f(\mathbf{x}_i^\top \boldsymbol{\beta}^*)] \cdot f'(\mathbf{x}_i^\top \boldsymbol{\beta}^*) \cdot \mathbf{x}_i = -\frac{1}{n}\sum_{i=1}^n f'(\mathbf{x}_i^\top \boldsymbol{\beta}^*) \cdot \mathbf{x}_i \cdot \epsilon_i. \quad (6.1)$$

Since $f'$ is bounded and that $\epsilon_i$'s are i.i.d. sub-Gaussian random variables, conditioning on $\{\mathbf{x}_i\}_{i=1}^n$, $\nabla L(\boldsymbol{\beta}^*)$ is the mean of i.i.d. sub-Gaussian random variables (Vershynin, 2010). Concentration of measure guarantees that $\nabla L(\boldsymbol{\beta}^*)$ is not far away from its mean, which is $\mathbf{0}$.

*Proof of Theorem 1.* For any stationary point $\widehat{\boldsymbol{\beta}}$ of the optimization problem in (1.2), by definition we have $\nabla L(\widehat{\boldsymbol{\beta}}) + \lambda \cdot \boldsymbol{\xi} = \mathbf{0}$, where $\boldsymbol{\xi} \in \partial \|\widehat{\boldsymbol{\beta}}\|_1$. For notational simplicity, we denote $\widehat{\boldsymbol{\beta}} - \boldsymbol{\beta}^*$ as $\boldsymbol{\delta}$. By definition, we have $\langle \nabla L(\widehat{\boldsymbol{\beta}}) - \nabla L(\boldsymbol{\beta}^*), \boldsymbol{\beta} - \boldsymbol{\beta}^* \rangle = \langle -\lambda \cdot \boldsymbol{\xi} - \nabla L(\boldsymbol{\beta}^*), \boldsymbol{\delta} \rangle$. We denote the support of $\boldsymbol{\beta}^*$ as $\mathcal{S}$, that is, $\mathcal{S} = \{j\colon \beta_j^* \neq 0\}$. By writing $\boldsymbol{\xi} = \boldsymbol{\xi}_\mathcal{S} + \boldsymbol{\xi}_{\mathcal{S}^c}$ we have

$$\langle \nabla L(\widehat{\boldsymbol{\beta}}) - \nabla L(\boldsymbol{\beta}^*), \widehat{\boldsymbol{\beta}} - \boldsymbol{\beta}^* \rangle = \langle -\lambda \cdot \boldsymbol{\xi}_{\mathcal{S}^c} - \lambda \cdot \boldsymbol{\xi}_\mathcal{S} - \nabla L(\boldsymbol{\beta}^*), \boldsymbol{\delta} \rangle. \quad (6.2)$$

Note that $\boldsymbol{\beta}_\mathcal{S}^* = \mathbf{0}$ and $\langle \boldsymbol{\xi}_{\mathcal{S}^c}, \widehat{\boldsymbol{\beta}}_{\mathcal{S}^c} \rangle = \|\widehat{\boldsymbol{\beta}}_{\mathcal{S}^c}\|_1 = \|\boldsymbol{\delta}_{\mathcal{S}^c}\|_1$. By Hölder's inequality, since $\|\boldsymbol{\xi}\|_\infty \leq 1$, the right-hand side of (6.2) can be bounded by

$$\langle \nabla L(\widehat{\boldsymbol{\beta}}) - \nabla L(\boldsymbol{\beta}^*), \widehat{\boldsymbol{\beta}} - \boldsymbol{\beta}^* \rangle \leq -\lambda \|\boldsymbol{\delta}_{\mathcal{S}^c}\|_1 + \lambda \|\boldsymbol{\delta}_\mathcal{S}\|_1 + \|\nabla L(\boldsymbol{\beta})\|_\infty \|\boldsymbol{\delta}\|_1. \quad (6.3)$$

Now we invoke Condition `Bounded-Gradient`$(B, \delta)$ to bound the right hand side of (6.3). We claim that Condition `Bounded-Gradient`$(B, \delta)$ holds with $\delta \leq (2d)^{-1}$, which will be verified in §A.1. In what follows, we condition on the event that $\|\nabla L(\boldsymbol{\beta}^*)\|_\infty \leq B\sigma \cdot \sqrt{\log d/n}$. By the definition of $\lambda$ and and Condition `Bounded-Gradient`$(B, \delta)$, we have $\lambda \geq L_1 \cdot \|\nabla L(\boldsymbol{\beta})\|_\infty$ with probability at least $1 - \delta$. By (6.3) we have

$$\langle \nabla L(\widehat{\boldsymbol{\beta}}) - \nabla L(\boldsymbol{\beta}^*), \widehat{\boldsymbol{\beta}} - \boldsymbol{\beta}^* \rangle \leq -\lambda\|\boldsymbol{\delta}_{\mathcal{S}^c}\|_1 + \lambda\|\boldsymbol{\delta}_\mathcal{S}\|_1 + L_1^{-1}\lambda\|\boldsymbol{\delta}\|_1. \tag{6.4}$$

Table 1: Asymptotic inference for nonlinear compressed sensing: Type I errors and powers of the decorrelated score and Wald tests for $H_0\colon \beta_j^* = 0$ at the 0.05-significance level. To obtain the type-I error, we test $H_0\colon \beta_{11}^* = 0$, and for the powers, we test $H_0\colon \beta_1^* = 0$. $\mu$ is the value of the nonzeros entries of $\boldsymbol{\beta}^*$.

|  | | Decorrelated Score Test | | Decorrelated Wald Test | |
| --- | --- | --- | --- | --- | --- |
|  | $\mu$ | Type-I error | Power | Type-I error | Power |
| 1 | 0.00 | 0.048 | 0.048 | 0.054 | 0.054 |
| 2 | 0.05 | 0.052 | 0.094 | 0.052 | 0.106 |
| 3 | 0.10 | 0.052 | 0.208 | 0.050 | 0.214 |
| 4 | 0.15 | 0.056 | 0.312 | 0.052 | 0.308 |
| 5 | 0.20 | 0.054 | 0.484 | 0.054 | 0.460 |
| 6 | 0.25 | 0.046 | 0.648 | 0.050 | 0.628 |
| 7 | 0.30 | 0.050 | 0.760 | 0.056 | 0.750 |
| 8 | 0.35 | 0.048 | 0.926 | 0.048 | 0.894 |
| 9 | 0.40 | 0.050 | 0.986 | 0.050 | 0.982 |
| 10 | 0.45 | 0.054 | 0.988 | 0.052 | 0.986 |
| 11 | 0.50 | 0.054 | 0.990 | 0.048 | 0.986 |

The next lemma, proven in §A.1, establishes a lower-bound of the left-hand side of (6.4).

**Lemma 6.** Recall that $\widehat{\boldsymbol{\Sigma}} := n^{-1} \sum_{i=1}^n \mathbf{x}_i \mathbf{x}_i^\top$. Under the Assumption `Bounded-Design`$(D)$, it holds with probability at least $1 - (2d)^{-1}$ that, for any $\boldsymbol{\beta} \in \mathbb{R}^d$,

$$\langle \nabla L(\boldsymbol{\beta}) - \nabla L(\boldsymbol{\beta}^*), \boldsymbol{\beta} - \boldsymbol{\beta}^* \rangle \geq a^2 (\boldsymbol{\beta} - \boldsymbol{\beta}^*)^\top \widehat{\boldsymbol{\Sigma}} (\boldsymbol{\beta} - \boldsymbol{\beta}^*) - 3b\sigma\sqrt{D \log d / n}\|\boldsymbol{\beta} - \boldsymbol{\beta}^*\|_1.$$

Combining (3.3), (6.4) and Lemma 6 we obtain that

$$0 \leq a^2 \boldsymbol{\delta}^\top \widehat{\boldsymbol{\Sigma}} \boldsymbol{\delta} \leq -\lambda\|\boldsymbol{\delta}_{\mathcal{S}^c}\|_1 + \lambda\|\boldsymbol{\delta}_\mathcal{S}\|_1 + \mu\lambda\|\boldsymbol{\delta}\|_1 = -\lambda(1-\mu)\|\boldsymbol{\delta}_{\mathcal{S}^c}\|_1 + \lambda(1+\mu)\|\boldsymbol{\delta}_\mathcal{S}\|_1. \tag{6.5}$$

where $\mu = L_1^{-1} + 3b\sqrt{D}L_2^{-1} \leq 0.1$. Hence it follows that $\|\boldsymbol{\delta}_{\mathcal{S}^c}\|_1 \leq (1+\mu)/(1-\mu)\|\boldsymbol{\delta}_\mathcal{S}\|_1 \leq 1.23\|\boldsymbol{\delta}_\mathcal{S}\|_1$.

Now we invoke the following lemma to bound $\boldsymbol{\delta}^\top \widehat{\boldsymbol{\Sigma}} \boldsymbol{\delta}$ from below.

**Lemma 7.** For any $\boldsymbol{\eta} \in \mathbb{R}^d$ and any index set $\mathcal{S}$ with $|\mathcal{S}| = s^*$, let $\mathcal{J}$ be the set of indices of the largest $k^*$ entries of $\boldsymbol{\eta}_{\mathcal{S}^c}$ in absolute value and let $\mathcal{I} = \mathcal{J} \cup \mathcal{S}$. Here $s^*$ and $k^*$ are the same as those in Assumption `Sparse-Eigenvalue`$(s^*, k^*)$. Assume that $\|\boldsymbol{\eta}_{\mathcal{S}^c}\|_1 \leq \gamma\|\boldsymbol{\eta}_\mathcal{S}\|_1$ for some $\gamma > 0$. Then we obtain that $\|\boldsymbol{\eta}\|_2 \leq (1+\gamma)\|\boldsymbol{\eta}_\mathcal{I}\|_2$ and that

$$\boldsymbol{\eta}^\top \widehat{\boldsymbol{\Sigma}} \boldsymbol{\eta} \geq \rho_-(s^* + k^*) \cdot \left[\|\boldsymbol{\eta}_\mathcal{I}\|_2 - \gamma\sqrt{s^*/k^*}\sqrt{\rho_+(k^*)/\rho_-(s^* + 2k^*) - 1} \cdot \|\boldsymbol{\eta}_\mathcal{S}\|_2\right] \cdot \|\boldsymbol{\eta}_\mathcal{I}\|_2. \tag{6.6}$$

By Assumption `Sparse-Eigenvalue`$(s^*, k^*)$ we have $\rho_+(k^*)/\rho_-(s^* + 2k^*) \leq 1 + 0.5k^*/s^*$. Combining this inequality with Lemma 7 we obtain that

$$\boldsymbol{\delta}^\top \widehat{\boldsymbol{\Sigma}} \boldsymbol{\delta} \geq \left(1 - 1.23\sqrt{0.5}\right) \cdot \rho_-(s^* + k^*) \cdot \|\boldsymbol{\delta}_\mathcal{I}\|_2^2 \geq 0.1 \cdot \rho_-(s^* + k^*) \cdot \|\boldsymbol{\delta}_\mathcal{I}\|_2^2, \tag{6.7}$$

where $\mathcal{J}$ is the set of indices of the largest $k^*$ entries of $\boldsymbol{\delta}_{\mathcal{S}^c}$ in absolute value and $\mathcal{I} = \mathcal{J} \cup \mathcal{S}$. Here the first inequality of (6.7) follows from Lemma 7 and that $\mathcal{S} \subset \mathcal{I}$. Combining (6.5) and (6.7) we obtain that

$$0.1 \cdot \rho_-(s^* + k^*) \cdot \|\boldsymbol{\delta}_{\mathcal{I}}\|_2^2 \leq \boldsymbol{\delta}^\top \widehat{\boldsymbol{\Sigma}} \boldsymbol{\delta} \leq a^{-2}\lambda(1+\mu)\|\boldsymbol{\delta}_{\mathcal{S}}\|_1 \leq 1.1 \cdot a^{-2}\sqrt{s^*}\lambda\|\boldsymbol{\delta}_{\mathcal{I}}\|_2,$$

which implies that $\|\boldsymbol{\delta}_{\mathcal{I}}\|_2 \leq 11/\rho_-(s^* + k^*) \cdot a^{-2}\sqrt{s^*}\lambda$. Note that by Lemma 7 we also have $\|\boldsymbol{\delta}\|_2 \leq 2.23\|\boldsymbol{\delta}_{\mathcal{I}}\|_2$. Combining this inequality with the fact that $\|\boldsymbol{\delta}_{\mathcal{S}^c}\|_1 \leq 1.23\|\boldsymbol{\delta}_{\mathcal{S}}\|_1$, we have

$$\|\widehat{\boldsymbol{\beta}} - \boldsymbol{\beta}^*\|_1 = \|\boldsymbol{\delta}\|_1 \leq 2.23\|\boldsymbol{\delta}_{\mathcal{S}}\|_1 \leq 2.23\sqrt{s^*}\|\boldsymbol{\delta}_{\mathcal{S}}\|_2 \leq 25/\rho_-(s^* + k^*) \cdot a^{-2}s^*\lambda \text{ and}$$
$$\|\widehat{\boldsymbol{\beta}} - \boldsymbol{\beta}^*\|_2 = \|\boldsymbol{\delta}\|_2 \leq 2.23\|\boldsymbol{\delta}_{\mathcal{I}}\|_2 \leq 25/\rho_-(s^* + k^*) \cdot a^{-2}\sqrt{s^*}\lambda.$$

Finally, to show that Algorithm 1 indeed catches a stationary point, we note that the acceptance criterion of the Algorithm (Line 10) implies that $\phi(\boldsymbol{\beta}^{(1)}) \leq \phi(\boldsymbol{\beta}^{(0)})$ where $\phi(\boldsymbol{\beta}) = L(\boldsymbol{\beta}) + \lambda\|\boldsymbol{\beta}\|_1$. Moreover, for $t = 2$, we also have $\phi(\boldsymbol{\beta}^{(2)}) \leq \max\{\phi(\boldsymbol{\beta}^{(0)}), \phi(\boldsymbol{\beta}^{(1)})\}$. By induction, we conclude that for all $t \geq 1$, $\phi(\boldsymbol{\beta}^{(t)}) \leq \phi(\boldsymbol{\beta}^{(0)})$. Therefore we have $\boldsymbol{\beta}^{(t)} \in \mathcal{C} := \{\boldsymbol{\beta} \in \mathbb{R}^d : \|\boldsymbol{\beta}\|_1 \leq \lambda^{-1} \cdot L(\boldsymbol{\beta}^{(0)}) + \|\boldsymbol{\beta}^{(0)}\|_1\}$. Since set $\mathcal{C}$ is compact and the loss function $L$ is continuously differentiable, it is also Lipschitz on $\mathcal{C}$. Therefore, by the convergence result of in Theorem 1 of Wright et al. (2009), we conclude that every accumulation point of Algorithm 1 is a stationary point of optimization problem (1.2). □

## 6.2 Proof of Theorem 2

In what follows, inspired by Raskutti et al. (2011), we apply Fano's method to derive the minimax risk of estimation for the nonlinear regression model defined in (1.1).

*Proof.* Let $M = M(\delta_n)$ be the cardinality of a $2\delta_n$-packing set of $B_0(s)$ with respect to the $\ell_2$-metric where $\delta_n$ will be specified later. We denote the elements of this packing set as $\{\boldsymbol{\beta}^1, \ldots, \boldsymbol{\beta}^M\}$. For any estimator $\widehat{\boldsymbol{\beta}}$, let $\psi = \operatorname{argmin}_{i \leq M} \|\widehat{\boldsymbol{\beta}} - \boldsymbol{\beta}^i\|_2$, triangle inequality implies that

$$2\|\widehat{\boldsymbol{\beta}} - \boldsymbol{\beta}^i\|_2 \geq \|\widehat{\boldsymbol{\beta}} - \boldsymbol{\beta}^i\|_2 + \|\widehat{\boldsymbol{\beta}} - \boldsymbol{\beta}^\psi\|_2 \geq \|\boldsymbol{\beta}^i - \boldsymbol{\beta}^\psi\|_2 \geq 2\delta_n \text{ for } i \neq \psi.$$

Thus we conclude that

$$\mathcal{R}_f^*(s, n, d) \geq \inf_\psi \sup_{1 \leq i \leq M} \delta_n^2 \cdot \mathbb{P}_{\boldsymbol{\beta}^i}(\psi \neq i) \geq \inf_\psi \delta_n^2 \cdot \mathbb{P}_{\boldsymbol{\beta}^U}(\psi \neq U),$$

where $U$ is uniform distributed over $\{1, \ldots, N\}$. We consider the following data-generating process: For a continuously differentiable function $f$ with $f'(u) \in [a, b], \forall u \in \mathbb{R}$, we first sample a random variable $U$ uniformly over $1, \ldots M$, then generate data $y_i = f(\mathbf{x}_i^\top \boldsymbol{\beta}^U) + \epsilon_i$. Fano's inequality implies that

$$\mathbb{P}(\psi \neq U) \geq 1 - \big[I(U; y_1, \ldots, y_n) + \log 2\big]/\log N.$$

In what follows, we establish an upper bound for the mutual information $I(U; y_1, \ldots, y_n)$. For $s \in \{1, \ldots, d\}$, we define the high-dimensional sparse hypercube as $\mathcal{C}_o(s) := \{\mathbf{v} \in \{0,1\}^d, \|\mathbf{v}\|_0 = s\}$. We define the Hamming distance on $\mathcal{C}_0(s)$ as $\rho_H(\mathbf{v}, \mathbf{v}') = \sum_{i=1}^d \mathbb{1}\{v_i \neq v_i'\}$. The following lemma, obtained from Rigollet et al. (2011), is an extension of the Varhsamov-Gilbert lemma to $\mathcal{C}_0(s)$.

**Lemma 8** (Sparse Varshamov-Gilbert lemma)**.** For any two integers $s$ and $d$ satisfying $1 \leq s \leq d/8$, there exist $\mathbf{v}_1, \ldots, \mathbf{v}_M \in \{0,1\}^d$ such that

$$\rho_H(\mathbf{v}_i, \mathbf{v}_j) \geq s/2 \text{ for all } i \neq j, \ \log(M) \geq s/8 \cdot \log[1 + d/(2s)], \text{ and } \|\mathbf{v}_i\|_0 = s \text{ for all } i.$$

By Lemma (8) there exist $\mathcal{C}' \subset \mathcal{C}_0$ with $|\mathcal{C}'| \geq \exp\{s/8 \cdot \log[1+d/(2s)]\}$ such that $\rho_H(\mathbf{v}, \mathbf{v}') \geq s/2$ for all $\mathbf{v}, \mathbf{v}' \in \mathcal{C}'$. Then for $\boldsymbol{\beta}, \boldsymbol{\beta}' \in \mathcal{C} := \delta_n \cdot \sqrt{2/s} \cdot \mathcal{C}'$, we have

$$\delta_n^2 \cdot 2/s \cdot \rho_H(\boldsymbol{\beta}, \boldsymbol{\beta}') \leq \|\boldsymbol{\beta} - \boldsymbol{\beta}'\|_2^2 \leq 2(\|\boldsymbol{\beta}\|_2^2 + \|\boldsymbol{\beta}'\|_2^2) \leq 8\delta_n^2,$$

which implies that $\delta_n^2 \leq \|\boldsymbol{\beta} - \boldsymbol{\beta}'\|_2^2 \leq 8\delta_n^2$ for all $\boldsymbol{\beta}, \boldsymbol{\beta}' \in \mathcal{C}$. By the convexity of mutual information, we have $I(U; y_1, \ldots, y_n) \leq M^{-2} \sum_{1 \leq m, m' \leq M} D_{KL}(\boldsymbol{\beta}^m, \boldsymbol{\beta}^{m'})$. Since given $\boldsymbol{\beta}$ and $f$, $y_i \sim N(f(\mathbf{x}_i^\top \boldsymbol{\beta}), \sigma^2)$, direct computation yields that

$$D_{KL}(\boldsymbol{\beta}^m, \boldsymbol{\beta}^{m'}) = 1/(2\sigma^2) \sum_{i=1}^n [f(\mathbf{x}_i^\top \boldsymbol{\beta}^m) - f(\mathbf{x}_i^\top \boldsymbol{\beta}^{m'})]. \tag{6.8}$$

By mean-value theorem, (6.8) can be bounded by

$$D_{KL}(\boldsymbol{\beta}^m, \boldsymbol{\beta}^{m'}) \leq n \cdot b^2/(2\sigma^2)(\boldsymbol{\beta}^m - \boldsymbol{\beta}^{m'})^\top \widehat{\boldsymbol{\Sigma}} (\boldsymbol{\beta}^m - \boldsymbol{\beta}^{m'})$$
$$\leq n \cdot b^2 \cdot \rho_+(2s)/(2\sigma^2) \|\boldsymbol{\beta}^m - \boldsymbol{\beta}^{m'}\|_2^2 \leq 4nb^2 \cdot \rho_+(2s) \cdot \delta_n^2/\sigma^2,$$

where the second inequality follows from $\|\boldsymbol{\beta}^m - \boldsymbol{\beta}^{m'}\|_0 \leq 2s$. Therefore we conclude that $I(U; y_1, \ldots, y_n) \leq 4nb^2 \cdot \rho_+(2s) \cdot \delta_n^2/\sigma^2$, which yields that

$$\inf_\psi \mathbb{P}_{\boldsymbol{\beta}}(\psi \neq U) \geq 1 - \frac{4nb^2 \cdot \rho_+(2s) \cdot \delta_n^2/\sigma^2 + \log 2}{\log M} \geq 1 - \frac{4nb^2 \cdot \rho_+(2s) \cdot \delta_n^2/\sigma^2 + \log 2}{s/8 \cdot \log[1 + d/(2s)]}.$$

Setting $\delta_n^2 = \frac{\sigma^2 s \log[1+d/(2s)]}{96nb^2 \rho_+(2s)}$, since $s \geq 4$ and $d \geq 8s$, we conclude that the right-hand side is no less than $1/2$. Now we obtain the following minimax lower bound

$$\mathcal{R}_f^*(s, n, d) \geq \frac{\sigma^2}{192 b^2 \rho_+(2s)} \frac{s \log[1 + d/(2s)]}{n}.$$

$\square$

## 6.3 Proof of Theorem 3

In the rest of this section, we prove our inferential results. Before delving into the proof, we first list two conditions that can simplify our introduction. We will verify these two conditions in §A.2. The first condition states that the Hessian of $L(\boldsymbol{\beta})$ is stable at the true parameter $\boldsymbol{\beta}^*$.

**Condition 9** . `Hessian-Stability`. For any $\boldsymbol{\beta} \in \mathbb{R}^d$, under Assumption 3, we have

$$\|\nabla^2 L(\boldsymbol{\beta}) - \nabla^2 L(\boldsymbol{\beta}^*)\|_\infty = \mathcal{O}_\mathbb{P}(\|\boldsymbol{\beta} - \boldsymbol{\beta}^*\|_1).$$

This condition characterizes the behavior of $\nabla^2 L$ at the true parameter $\boldsymbol{\beta}^*$. It states that $\nabla^2 L(\boldsymbol{\beta})$ deviates from $\nabla^2 L(\boldsymbol{\beta}^*)$ in the order of $\|\boldsymbol{\beta} - \boldsymbol{\beta}^*\|_1$.

The next condition quantifies the rate of $\nabla^2 L(\boldsymbol{\beta}^*)$ converging to its expectation.

**Condition 10** . `Hessian-Concentration`($C^h$). For the true parameter $\boldsymbol{\beta}^*$, it holds that

$$\|\nabla^2 L(\boldsymbol{\beta}^*) - \mathbb{E}[\nabla^2 L(\boldsymbol{\beta}^*)]\|_\infty \leq C^h \cdot \sqrt{\log d/n}$$

with probability tending to one for some $C^h > 0$.

Now we begin to prove the main theorem of the decorrelated score test.

*Proof of Theorem 3.* We first show that $\sqrt{n}\bigl|F_S(\widetilde{\boldsymbol{\beta}},\rho) - S(\boldsymbol{\beta}^*;\mathbf{d}^*)\bigr| = o_\mathbb{P}(1)$ where $S(\boldsymbol{\beta}^*;\mathbf{d}^*) := \nabla_\alpha L(\boldsymbol{\beta}^*) - \mathbf{d}^{*\top}\nabla_\gamma L(\boldsymbol{\beta}^*)$. Under the null hypothesis $H_0\colon \alpha^* = 0$, by Taylor expansion, there exists a $\boldsymbol{\beta}_1 = (0, \widetilde{\boldsymbol{\gamma}}^\top)^\top$ with $\widetilde{\boldsymbol{\gamma}}$ in the line segment between $\widehat{\boldsymbol{\gamma}}$ and $\boldsymbol{\gamma}^*$ such that,

$$\bigl|F_S(\widetilde{\boldsymbol{\beta}},\rho) - S(\boldsymbol{\beta}^*;\mathbf{d}^*)\bigr| \leq \bigl[\nabla_{\alpha\gamma}L(\boldsymbol{\beta}_1) - \widehat{\mathbf{d}}^\top\nabla_{\gamma\gamma}L(\boldsymbol{\beta}_1)\bigr](\widetilde{\boldsymbol{\beta}} - \boldsymbol{\beta}^*) + (\widehat{\mathbf{d}} - \mathbf{d}^*)^\top\nabla_\gamma L(\boldsymbol{\beta}^*),$$

where we denote $\widehat{\mathbf{d}} = \boldsymbol{d}(\widetilde{\boldsymbol{\beta}},\rho)$ for notational simplicity. Hölder's inequality implies that

$$\bigl|F_S(\widetilde{\boldsymbol{\beta}},\rho) - S(\boldsymbol{\beta}^*;\mathbf{d}^*)\bigr| \leq \bigl\|\nabla_{\alpha\gamma}L(\boldsymbol{\beta}_1) - \widehat{\mathbf{d}}^\top\nabla_{\gamma\gamma}L(\boldsymbol{\beta}_1)\bigr\|_\infty\bigl\|\widetilde{\boldsymbol{\beta}} - \boldsymbol{\beta}^*\bigr\|_1 + \bigl\|\widehat{\mathbf{d}} - \mathbf{d}^*\bigr\|_1\bigl\|\nabla L(\boldsymbol{\beta}^*)\bigr\|_\infty. \tag{6.9}$$

By triangle inequality, we have

$$\bigl\|\nabla_{\gamma\alpha}L(\boldsymbol{\beta}_1) - \widehat{\mathbf{d}}^\top\nabla_{\gamma\gamma}L(\boldsymbol{\beta}_1)\bigr\|_\infty \leq \bigl|\nabla_{\gamma\alpha}L(\boldsymbol{\beta}_1) - \nabla_{\gamma\alpha}L(\widetilde{\boldsymbol{\beta}})\bigr| + \|\widehat{\mathbf{d}}\|_1\cdot\bigl\|\nabla_{\gamma\gamma}L(\boldsymbol{\beta}_1) - \nabla_{\gamma\gamma}L(\widetilde{\boldsymbol{\beta}})\bigr\|_\infty$$
$$+ \bigl\|\nabla_{\alpha\gamma}L(\widetilde{\boldsymbol{\beta}}) - \widehat{\mathbf{d}}^\top\nabla_{\gamma\gamma}L(\widetilde{\boldsymbol{\beta}})\bigr\|_\infty. \tag{6.10}$$

By the definition of the Dantzig-selector (2.7), we have $\bigl\|\nabla_{\alpha\gamma}L(\widetilde{\boldsymbol{\beta}}) - \widehat{\mathbf{d}}^\top\nabla_{\gamma\gamma}L(\widetilde{\boldsymbol{\beta}})\bigr\|_\infty \leq \rho$. Hence by Hölder's inequality (6.10) is reduced to

$$\bigl\|\nabla_{\gamma\alpha}L(\boldsymbol{\beta}_1) - \widehat{\mathbf{d}}^\top\nabla_{\gamma\gamma}L(\boldsymbol{\beta}_1)\bigr\|_\infty \leq (1 + \|\widehat{\mathbf{d}}\|_1)\|\nabla^2 L(\boldsymbol{\beta}_1) - \nabla^2 L(\widehat{\boldsymbol{\beta}}')\|_\infty + \rho.$$

Under Condition `Hessian-Stability`, we have

$$\bigl\|\nabla^2 L(\boldsymbol{\beta}_1) - \nabla^2 L(\widehat{\boldsymbol{\beta}}')\bigr\|_\infty \leq \bigl\|\nabla^2 L(\widehat{\boldsymbol{\beta}}') - \nabla^2 L(\boldsymbol{\beta}^*)\bigr\|_\infty + \bigl\|\nabla^2 L(\boldsymbol{\beta}_1) - \nabla^2 L(\boldsymbol{\beta}^*)\bigr\|_\infty$$
$$= \mathcal{O}_\mathbb{P}(\|\widetilde{\boldsymbol{\beta}} - \boldsymbol{\beta}^*\|_1 + \|\boldsymbol{\beta}_1 - \boldsymbol{\beta}^*\|_1) = \mathcal{O}_\mathbb{P}(s^*\lambda),$$

where the last equation follows from Theorem 1. Hence we conclude that

$$\bigl\|\nabla_{\gamma\alpha}L(\boldsymbol{\beta}_1) - \widehat{\mathbf{d}}^\top\nabla_{\gamma\gamma}L(\boldsymbol{\beta}_1)\bigr\|_\infty \leq \mathcal{O}_\mathbb{P}\bigl[(1 + \|\widehat{\mathbf{d}}\|_1)\cdot s^*\lambda + \rho\bigr]. \tag{6.11}$$

Recalling Condition `Bounded-Gradient`$(B, \delta)$, we have $\bigl\|\nabla L(\boldsymbol{\beta}^*)\bigr\|_\infty = \mathcal{O}_\mathbb{P}(\sqrt{\log d/n}) = \mathcal{O}_\mathbb{P}(\lambda)$. Thus to bound $\bigl|F_S(\widetilde{\boldsymbol{\beta}};\rho) - S(\boldsymbol{\beta}^*;\mathbf{d}^*)\bigr|$, we only need to control $\|\widehat{\mathbf{d}} - \mathbf{d}^*\|_1$ and $\|\widehat{\mathbf{d}}\|_1$. The following lemma characterizes the statistical accuracy of $\widehat{\mathbf{d}}$.

**Lemma 11.** *Under Assumptions 3 and 4, $\mathbf{d}^*$ defined in (3.7) is feasible for the Dantzig selector problem in (2.7) with high probability. Moreover, for $\mathbf{d}(\boldsymbol{\beta},\rho)$ defined in (2.7), it holds with high probability that $\max\bigl\{\|\mathbf{d}(\widehat{\boldsymbol{\beta}},\rho) - \mathbf{d}^*\|_1, \|\mathbf{d}(\widetilde{\boldsymbol{\beta}},\rho) - \mathbf{d}^*\|_1\bigr\} \leq 16/\tau_* \cdot \rho \cdot s_\mathbf{d}^*$.*

In what follows, our analysis conditions on the event that $\mathbf{d}^*$ is feasible for (2.7) and that $\|\widehat{\mathbf{d}} - \mathbf{d}^*\|_1 \leq 16/\tau_* \cdot \rho \cdot s_\mathbf{d}^*$, which holds with probability tending to one. The optimality of $\widehat{\mathbf{d}}$ implies that $\|\widehat{\mathbf{d}}\|_1 \leq \|\mathbf{d}^*\|_1$. Therefore combining (6.9), (6.11) and Lemma 11 we conclude that

$$\bigl|F_S(\widetilde{\boldsymbol{\beta}};\rho) - S(\boldsymbol{\beta}^*;\mathbf{d}^*)\bigr| = \bigl\{\bigl[(1 + \|\mathbf{d}^*\|_1)\cdot s^*\lambda + \rho\bigr]\cdot s^*\lambda + s_\mathbf{d}^*\cdot\lambda\cdot\rho\bigr\}$$
$$= \mathcal{O}_\mathbb{P}\bigl[(s^* + s_\mathbf{d}^*)\cdot\lambda\cdot\rho\bigr] = o_\mathbb{P}(n^{-1/2}).$$

In the rest of the proof we show that $\widehat{\sigma}_S^2(\widetilde{\boldsymbol{\beta}},\rho)$ is a consistent estimator of the variance of $\sqrt{n}\cdot S(\boldsymbol{\beta}^*,\mathbf{d}^*)$, denoted as $\sigma_S^2(\boldsymbol{\beta}^*)$. By simple computation, we have $\sigma_S^2(\boldsymbol{\beta}^*) = \sigma_0^2 \cdot \sigma^2(\boldsymbol{\beta}^*)$ where

$$\mathbb{E}(\epsilon^2) = \sigma_0^2 \quad \text{and} \quad \sigma^2(\boldsymbol{\beta}^*) = [1, -(\mathbf{d}^*)^\top] \cdot \mathbf{I}(\boldsymbol{\beta}^*) \cdot [1, -(\mathbf{d}^*)^\top]^\top, \tag{6.12}$$

For notational simplicity, we denote $\widehat{\boldsymbol{\theta}} = [1, -(\widehat{\mathbf{d}})^\top]^\top$, $\boldsymbol{\theta} = [1, -(\mathbf{d}^*)^\top]^\top$ and define

$$\widehat{\sigma}_0^2(\boldsymbol{\beta}) = \frac{1}{n}\sum_{i=1}^n [y_i - f(\mathbf{x}_i^\top\boldsymbol{\beta})]^2 \quad \text{and} \quad \boldsymbol{\Lambda}(\boldsymbol{\beta}) := \frac{1}{n}\sum_{i=1}^n f'(\mathbf{x}_i^\top\boldsymbol{\beta})^2\mathbf{x}_i\mathbf{x}_i^\top. \tag{6.13}$$

By the definition of $\mathbf{\Lambda}(\boldsymbol{\beta})$, we have $\mathbb{E}[\mathbf{\Lambda}(\boldsymbol{\beta}^*)] = \mathbf{I}(\boldsymbol{\beta}^*)$ and $\operatorname{Var}(S(\boldsymbol{\beta}^*; \mathbf{d}^*)) = \sigma_0^2 \cdot \boldsymbol{\theta}^\top \mathbb{E}[\mathbf{\Lambda}(\boldsymbol{\beta}^*)] \boldsymbol{\theta}$. In addition, we denote

$$\widehat{\sigma}^2(\boldsymbol{\beta}, \rho) = [1, -\mathbf{d}(\boldsymbol{\beta}, \rho)^\top] \cdot \mathbf{\Lambda}(\boldsymbol{\beta}) [1, -\mathbf{d}(\boldsymbol{\beta}, \rho)^\top]^\top. \tag{6.14}$$

Hence by definition, $\widehat{\sigma}_S^2(\widetilde{\boldsymbol{\beta}}, \rho) = \widehat{\sigma}_0^2(\widetilde{\boldsymbol{\beta}}) \cdot \widehat{\sigma}^2(\widetilde{\boldsymbol{\beta}}, \rho)$. In the following we show that $\widehat{\sigma}_0^2(\widetilde{\boldsymbol{\beta}})$ and $\widehat{\sigma}^2(\widetilde{\boldsymbol{\beta}}, \rho)$ are consistent estimators of $\sigma_0^2$ and $\sigma^2(\boldsymbol{\beta}^*)$ respectively.

For $\widehat{\sigma}_0^2(\widetilde{\boldsymbol{\beta}})$, by $y_i = f(\mathbf{x}_i^\top \boldsymbol{\beta}^*) + \epsilon_i$ we have

$$\widehat{\sigma}_0^2(\widetilde{\boldsymbol{\beta}}) = \underbrace{\frac{1}{n} \sum_{i=1}^n \epsilon_i^2}_{(\mathrm{i}).a} + \underbrace{\frac{2}{n} \sum_{i=1}^n \epsilon_i \cdot [f(\mathbf{x}_i^\top \widetilde{\boldsymbol{\beta}}) - f(\mathbf{x}_i^\top \boldsymbol{\beta}^*)]}_{(\mathrm{i}).b} + \underbrace{\frac{1}{n} \sum_{i=1}^n [f(\mathbf{x}_i^\top \widetilde{\boldsymbol{\beta}}) - f(\mathbf{x}_i^\top \boldsymbol{\beta}^*)]^2}_{(\mathrm{i}).c}. \tag{6.15}$$

For term (i).$a$ in (6.15), strong law of large numbers implies that (i).$a \to \sigma_0^2$ almost surely. Recall that $\epsilon_i$ is a sub-Gaussian random variable with variance proxy $\sigma^2$. Thus term (i).$b$ in (6.15) is also sub-Gaussian with variance proxy bounded by

$$\frac{4\sigma^2}{n^2} \sum_{i=1}^n \bigl[f(\mathbf{x}_i^\top \widetilde{\boldsymbol{\beta}}) - f(\mathbf{x}_i^\top \boldsymbol{\beta}^*)\bigr]^2 = \frac{4\sigma^2}{n^2} \sum_{i=1}^n f'(\mathbf{x}_i^\top \widebar{\boldsymbol{\beta}})^2 \bigl[\mathbf{x}_i^\top (\widetilde{\boldsymbol{\beta}} - \boldsymbol{\beta}^*)\bigr]^2. \tag{6.16}$$

where $\widebar{\boldsymbol{\beta}}$ is an intermediate value between $\widetilde{\boldsymbol{\beta}}$ and $\boldsymbol{\beta}^*$. Recall that $f'(u) \le b$ for any $u \in \mathbb{R}$ and we prove in §A.2. Assumption Bounded-Design($D$) holds with high probability since $\mathbf{x}$ is sub-Gaussian. By Hölder's inequality, the right-hand side of (6.16) is further bounded in high probability by

$$4\sigma^2 \cdot b^2 / n \cdot (\widetilde{\boldsymbol{\beta}} - \boldsymbol{\beta}^*)^\top \widehat{\boldsymbol{\Sigma}} (\widetilde{\boldsymbol{\beta}} - \boldsymbol{\beta}^*) \le 4\sigma^2 \cdot b^2 / n \|\widetilde{\boldsymbol{\beta}} - \boldsymbol{\beta}^*\|_1^2 \cdot \|\widehat{\boldsymbol{\Sigma}}\|_\infty = \mathcal{O}_{\mathbb{P}}\bigl[(s^*\lambda)^2 / n\bigr].$$

By the tail bound for sub-Gaussian random variable, we conclude that term (i).$b = \mathcal{O}_{\mathbb{P}}(s^* \lambda \cdot \sqrt{\log d / n}) = o_{\mathbb{P}}(1)$. Under Assumption Bounded-Design($D$), for term (i).$c$ in (6.15), mean-value theorem implies that

$$(\mathrm{i}).c = \sum_{i=1}^n f'(\mathbf{x}_i^\top \widebar{\boldsymbol{\beta}})^2 \bigl[\mathbf{x}_i^\top (\widetilde{\boldsymbol{\beta}} - \boldsymbol{\beta}^*)\bigr]^2 \le b^2 (\widetilde{\boldsymbol{\beta}} - \boldsymbol{\beta}^*)^\top \widehat{\boldsymbol{\Sigma}} (\widetilde{\boldsymbol{\beta}} - \boldsymbol{\beta}^*) = \mathcal{O}_{\mathbb{P}}\bigl[(s^*\lambda)^2\bigr] = o_{\mathbb{P}}(1).$$

Therefore, combining the upper bounds for each term on the right-hand side of (6.15), we conclude that $\widehat{\sigma}_0^2(\widetilde{\boldsymbol{\beta}})$ converges to $\sigma_0^2$ in probability.

It remains to show that $\widehat{\sigma}^2(\widetilde{\boldsymbol{\beta}}, \rho)$ is consistent for estimating $\sigma^2(\boldsymbol{\beta}^*)$. Triangle inequality implies that

$$\begin{aligned}
\bigl|\widehat{\sigma}(\widetilde{\boldsymbol{\beta}}, \rho) - \sigma^2(\boldsymbol{\beta}^*)\bigr| &= \bigl|\widehat{\boldsymbol{\theta}}^\top \mathbf{\Lambda}(\widetilde{\boldsymbol{\beta}}) \widehat{\boldsymbol{\theta}} - \boldsymbol{\theta}^\top \mathbb{E}[\mathbf{\Lambda}(\boldsymbol{\beta}^*)] \boldsymbol{\theta}\bigr| \\
&\le \underbrace{\bigl|\widehat{\boldsymbol{\theta}}^\top \{\mathbf{\Lambda}(\widetilde{\boldsymbol{\beta}}) - \mathbb{E}[\mathbf{\Lambda}(\boldsymbol{\beta}^*)]\} \widehat{\boldsymbol{\theta}}\bigr|}_{(\mathrm{ii}).a} + \underbrace{\bigl|\widehat{\boldsymbol{\theta}}^\top \mathbb{E}[\mathbf{\Lambda}(\boldsymbol{\beta}^*)] \widehat{\boldsymbol{\theta}} - \boldsymbol{\theta}^\top \mathbb{E}[\mathbf{\Lambda}(\boldsymbol{\beta}^*)] \boldsymbol{\theta}\bigr|}_{(\mathrm{ii}).b},
\end{aligned}$$

For term (ii).$a$, Hölder's inequality and the optimality of $\widehat{\mathbf{d}}$ implies that

$$(\mathrm{ii}).a \le \|\widehat{\boldsymbol{\theta}}\|_1^2 \cdot \|\mathbf{\Lambda}(\widetilde{\boldsymbol{\beta}}) - \mathbb{E}[\mathbf{\Lambda}(\boldsymbol{\beta}^*)]\|_\infty \le (1 + \|\mathbf{d}^*\|_1)^2 \cdot \|\mathbf{\Lambda}(\widetilde{\boldsymbol{\beta}}) - \mathbb{E}[\mathbf{\Lambda}(\boldsymbol{\beta}^*)]\|_\infty. \tag{6.17}$$

The following lemma gives a bound on the right-hand side of (6.17).

**Lemma 12.** We denote $\widehat{\boldsymbol{\beta}} = (\widehat{\alpha}, \widehat{\boldsymbol{\gamma}}^\top)^\top$ to be the attained $\ell_1$-regularized least-square estimator and $\widetilde{\boldsymbol{\beta}} = (0, \widehat{\boldsymbol{\gamma}}^\top)^\top$. Under Assumptions 3 and 4, for $\mathbf{\Lambda}(\boldsymbol{\beta}) \in \mathbb{R}^{d \times d}$ defined in (6.13), we have

$$\|\mathbf{\Lambda}(\widetilde{\boldsymbol{\beta}}) - \mathbb{E}[\mathbf{\Lambda}(\boldsymbol{\beta}^*)]\|_\infty = \mathcal{O}_{\mathbb{P}}(s^* \lambda).$$

Thus by (6.17) and Lemma (12) we conclude that (ii).$a \leq (1 + \|\mathbf{d}^*\|_1)^2 \cdot s^*\lambda$. Finally for term (ii).$b$, triangle inequality and Hölder's inequalty imply that

$$\text{(ii)}.b \leq |(\widehat{\boldsymbol{\theta}} - \boldsymbol{\theta})^\top \mathbb{E}[\boldsymbol{\Lambda}(\boldsymbol{\beta}^*)](\widehat{\boldsymbol{\theta}} - \boldsymbol{\theta})| + 2|\boldsymbol{\theta}^\top \mathbb{E}[\boldsymbol{\Lambda}(\boldsymbol{\beta}^*)](\widehat{\boldsymbol{\theta}} - \boldsymbol{\theta})|$$
$$\leq (\|\widehat{\boldsymbol{\theta}} - \boldsymbol{\theta}\|_1^2 + 2\|\boldsymbol{\theta}\|_1 \cdot \|\widehat{\boldsymbol{\theta}} - \boldsymbol{\theta}\|_1) \cdot \|\mathbb{E}[\boldsymbol{\Lambda}(\boldsymbol{\beta}^*)]\|_\infty.$$

Combining $\|\widehat{\boldsymbol{\theta}} - \boldsymbol{\theta}\|_1 = \|\widehat{\mathbf{d}} - \mathbf{d}^*\|_1 = \mathcal{O}_\mathbb{P}(s_{\mathbf{d}}^* \cdot \rho)$, $\|\boldsymbol{\theta}\|_1 = 1 + \|\mathbf{d}^*\|_1$ and $\|\mathbb{E}[\boldsymbol{\Lambda}(\boldsymbol{\beta}^*)]\|_\infty = \mathcal{O}_\mathbb{P}(1)$, we have

$$\text{(ii)}.b = \mathcal{O}_\mathbb{P}\big[(s_{\mathbf{d}}^* \cdot \rho)^2 + (1 + \|\mathbf{d}^*\|_1) \cdot s_{\mathbf{d}}^* \cdot \rho\big] = \mathcal{O}_\mathbb{P}\big[(1 + \|\mathbf{d}^*\|_1) \cdot s_{\mathbf{d}}^* \cdot \rho\big].$$

Thus we conclude that $\widehat{\sigma}(\widetilde{\boldsymbol{\beta}}; \rho)$ is a consistent estimator of $\sigma_S^2(\boldsymbol{\beta}^*)$. Together with the consistency of $\widehat{\sigma}_0^2(\widetilde{\boldsymbol{\beta}})$, we conclude that $\widehat{\sigma}_S^2(\widetilde{\boldsymbol{\beta}}, \rho)$ is a consistent estimator of the variance of $\sqrt{n} \cdot S(\boldsymbol{\beta}^*; \mathbf{d}^*)$. Finally, by invoking Slutsky's theorem, it holds that $\sqrt{n} \cdot F_S(\widetilde{\boldsymbol{\beta}}, \rho)/\widehat{\sigma}_S(\widetilde{\boldsymbol{\beta}}, \rho) \rightsquigarrow N(0, 1)$. □

## 6.4 Proof of Theorem 4

After proving Theorem 3, we are ready to prove Theorem 4, which shows the validity of the high-dimensional Wald test.

*Proof of Theorem 4.* We first show that $\sqrt{n} \cdot [\bar{\alpha}(\widehat{\boldsymbol{\beta}}, \rho) - \alpha^*] \rightsquigarrow N\big(0, \sigma_0^2/\sigma^2(\boldsymbol{\beta}^*)\big)$ where $\sigma_0^2 = \mathbb{E}(\epsilon^2)$ and $\sigma^2(\boldsymbol{\beta}^*)$ is defined in (6.12). For notational simplicity, for $\boldsymbol{\beta} = (\alpha, \boldsymbol{\gamma}^\top)^\top$ and $\mathbf{v} \in \mathbb{R}^{d-1}$ we define $S(\boldsymbol{\beta}; \mathbf{v}) := \nabla_\alpha L(\boldsymbol{\beta}) - \mathbf{v}^\top \nabla_{\boldsymbol{\gamma}} L(\boldsymbol{\beta})$. For $\mathbf{d}(\cdot, \cdot)$ defined in (2.7), we denote $\mathbf{d}(\widehat{\boldsymbol{\beta}}, \rho)$ as $\widehat{\mathbf{d}}$ where $\widehat{\boldsymbol{\beta}} = (\widehat{\alpha}, \widehat{\boldsymbol{\gamma}}^\top)^\top$ is the $\ell_1$-regularized least-square estimator. By definition, we have $S(\widehat{\boldsymbol{\beta}}; \widehat{\mathbf{d}}) = F_S(\widehat{\boldsymbol{\beta}}, \rho)$ where the decorrelated score function $F_S(\cdot, \cdot)$ is defined in (2.6). Let $\widetilde{\boldsymbol{\beta}} = (\alpha^*, \widehat{\boldsymbol{\gamma}}^\top)^\top$. Taylor's expansion of $S(\cdot; \widehat{\mathbf{d}})$ yields that

$$S(\widehat{\boldsymbol{\beta}}; \widehat{\mathbf{d}}) = S(\boldsymbol{\beta}^*; \widehat{\mathbf{d}}) + (\widehat{\alpha} - \alpha^*) \cdot \nabla_\alpha S(\breve{\boldsymbol{\beta}}; \widehat{\mathbf{d}}), \tag{6.18}$$

where $\breve{\boldsymbol{\beta}}$ is an intermediate value between $\widehat{\boldsymbol{\beta}}$ and $\widetilde{\boldsymbol{\beta}}$, and we have

$$\nabla_\alpha S(\boldsymbol{\beta}; \mathbf{v}) = \nabla_{\alpha\alpha}^2 L(\boldsymbol{\beta}) - \nabla_{\alpha\boldsymbol{\gamma}}^2 L(\boldsymbol{\beta}) \cdot \mathbf{v}. \tag{6.19}$$

Thus, by (6.20) and the definition in (2.12), we obtain that $\bar{\alpha}(\widehat{\boldsymbol{\beta}}, \rho) = \widehat{\alpha} - \big[\nabla_\alpha S(\widehat{\boldsymbol{\beta}}; \widehat{\mathbf{d}})\big]^{-1} \cdot S(\widehat{\boldsymbol{\beta}}; \widehat{\mathbf{d}})$. In addition, by (6.18) we can write $\sqrt{n} \cdot [\bar{\alpha}(\widehat{\boldsymbol{\beta}}, \rho) - \alpha^*]$ as

$$\sqrt{n} \cdot [\bar{\alpha}(\widehat{\boldsymbol{\beta}}, \rho) - \alpha^*] = \overbrace{\sqrt{n} \cdot \big[\nabla_\alpha S(\widehat{\boldsymbol{\beta}}; \widehat{\mathbf{d}})\big]^{-1} \cdot S(\boldsymbol{\beta}^*; \widehat{\mathbf{d}})}^{\text{(i)}}$$
$$+ \underbrace{\sqrt{n} \cdot (\widehat{\alpha} - \alpha^*) \cdot \Big\{1 - \big[\nabla_\alpha S(\widehat{\boldsymbol{\beta}}; \widehat{\mathbf{d}})\big]^{-1} \cdot \nabla_\alpha S(\breve{\boldsymbol{\beta}}; \widehat{\mathbf{d}})\Big\}}_{\text{(ii)}}. \tag{6.20}$$

For term (i) in (6.20), we show that

$$\nabla_\alpha S(\widehat{\boldsymbol{\beta}}; \widehat{\mathbf{d}}) = \sigma^2(\boldsymbol{\beta}^*) + o_\mathbb{P}(1), \tag{6.21}$$

where $\sigma^2(\boldsymbol{\beta}^*)$ is defined in (6.12). Note that by the definition of $\mathbf{d}^*$ we have

$$\sigma^2(\boldsymbol{\beta}^*) = \big\{[\mathbf{I}(\boldsymbol{\beta}^*)]_{\alpha,\alpha} - [\mathbf{I}(\boldsymbol{\beta}^*)]_{\alpha,\boldsymbol{\gamma}} \cdot \mathbf{d}^*\big\}.$$

Together with (6.19), by triangle inequality we further obtain

$$\big|\nabla_\alpha S(\widehat{\boldsymbol{\beta}}; \widehat{\mathbf{d}}) - \sigma^2(\boldsymbol{\beta}^*)\big| \leq \underbrace{\big|\nabla_{\alpha\alpha}^2 L(\widehat{\boldsymbol{\beta}}) - [\mathbf{I}(\boldsymbol{\beta}^*)]_{\alpha,\alpha}\big|}_{\text{(i)}.a} + \underbrace{\big|\nabla_{\alpha\boldsymbol{\gamma}}^2 L(\widehat{\boldsymbol{\beta}}) \cdot \widehat{\mathbf{d}} - [\mathbf{I}(\boldsymbol{\beta}^*)]_{\alpha,\boldsymbol{\gamma}} \cdot \mathbf{d}^*\big|}_{\text{(i)}.b}. \tag{6.22}$$

For term (i).$a$ in (6.22), triangle inequality implies the following upper bound for (i).$a$

$$\|\nabla^2 L(\widehat{\boldsymbol{\beta}}) - \mathbf{I}(\boldsymbol{\beta}^*)\|_\infty \leq \|\nabla^2 L(\widehat{\boldsymbol{\beta}}) - \nabla^2 L(\boldsymbol{\beta}^*)\|_\infty + \|\nabla^2 L(\boldsymbol{\beta}^*) - \mathbf{I}(\boldsymbol{\beta}^*)\|_\infty. \tag{6.23}$$

Combining Condition `Hessian-Stability`, Condition `Hessian-Concentration`($C^h$) and (6.23) together we conclude that

$$(i).a \leq \mathcal{O}_\mathbb{P}(\|\widehat{\boldsymbol{\beta}} - \boldsymbol{\beta}^*\|_1) + C^h\sqrt{\log d/n} = \mathcal{O}_\mathbb{P}(s^*\lambda) = o_\mathbb{P}(1). \tag{6.24}$$

For term (i).$b$ in (6.22), triangle inequality implies that

$$(i).b \leq \left|\left\{\nabla^2_{\alpha\gamma} L(\widehat{\boldsymbol{\beta}}) - [\mathbf{I}(\boldsymbol{\beta}^*)]_{\alpha,\gamma}\right\} \cdot (\widehat{\mathbf{d}} - \mathbf{d}^*)\right| + \left|[\mathbf{I}(\boldsymbol{\beta}^*)]_{\alpha,\gamma} \cdot (\mathbf{d}^* - \widehat{\mathbf{d}})\right| \\ + \left|\left\{\nabla^2_{\alpha\gamma} L(\widehat{\boldsymbol{\beta}}) - [\mathbf{I}(\boldsymbol{\beta}^*)]_{\alpha,\gamma}\right\} \cdot \mathbf{d}^*\right|. \tag{6.25}$$

Applying Hölder's inequality to each term on the right hand side of (6.25), we obtain that

$$(i).b \leq \|\widehat{\mathbf{d}} - \mathbf{d}^*\|_1 \cdot \|\nabla^2 L(\widehat{\boldsymbol{\beta}}) - \mathbf{I}(\boldsymbol{\beta}^*)\|_\infty + \|\widehat{\mathbf{d}} - \mathbf{d}^*\|_1 \cdot \|\mathbf{I}(\boldsymbol{\beta}^*)\|_\infty \\ + \|\mathbf{d}^*\|_1 \cdot \|\nabla^2 L(\widehat{\boldsymbol{\beta}}) - \mathbf{I}(\boldsymbol{\beta}^*)\|_\infty. \tag{6.26}$$

Combining Condition `Hessian-Stability`, Condition `Hessian-Concentration`($C^h$), Lemma 11, and Assumption 4 together we conclude that

$$(i).b \leq \mathcal{O}_\mathbb{P}(s^*_{\mathbf{d}}\rho \cdot s^*\lambda) + \mathcal{O}_\mathbb{P}(s^*_{\mathbf{d}} \cdot \rho) + \mathcal{O}_\mathbb{P}(\|\mathbf{d}^*\|_1 \cdot s^*\lambda) = \mathcal{O}_\mathbb{P}(s^*_{\mathbf{d}} \cdot \rho) = o_\mathbb{P}(1). \tag{6.27}$$

Therefore, combining (6.24) and (6.27) we establish that $\nabla_\alpha S(\widehat{\boldsymbol{\beta}}; \widehat{\mathbf{d}}) = \sigma^2(\boldsymbol{\beta}^*) + o_\mathbb{P}(1)$. Meanwhile, as shown in the in §6.3, $S(\widehat{\boldsymbol{\beta}}; \widehat{\mathbf{d}}) = F_S(\widehat{\boldsymbol{\beta}}, \rho) = S(\boldsymbol{\beta}^*; \mathbf{d}^*) + o_\mathbb{P}(n^{-1/2})$. By Slutsky's theorem and the asymptotic variance of $\sqrt{n} \cdot S(\boldsymbol{\beta}^*)$ derived in §6.3, this further implies that

$$\sqrt{n} \cdot \left[\nabla_\alpha S(\widehat{\boldsymbol{\beta}}; \widehat{\mathbf{d}})\right]^{-1} \cdot S(\widehat{\boldsymbol{\beta}}; \widehat{\mathbf{d}}) = \sqrt{n}/\sigma^2(\boldsymbol{\beta}^*) \cdot S(\boldsymbol{\beta}^*; \mathbf{d}^*) + o(1) \rightsquigarrow N(0, \sigma_0^2/\sigma^2(\boldsymbol{\beta}^*)).$$

For term (ii) in (6.20), we show that it is asymptotically negligible. First we have

$$(ii) \leq \sqrt{n} \cdot \underbrace{|\widehat{\alpha} - \alpha^*|}_{(ii).a} \cdot \underbrace{\left|1 - \left[\nabla_\alpha S(\widehat{\boldsymbol{\beta}}; \widehat{\mathbf{d}})\right]^{-1} \cdot \nabla_\alpha S(\widetilde{\boldsymbol{\beta}}; \widehat{\mathbf{d}})\right|}_{(ii).b}. \tag{6.28}$$

By Theorem 1 we have term $(ii).a \leq \|\widehat{\boldsymbol{\beta}} - \boldsymbol{\beta}^*\|_1 = \mathcal{O}_\mathbb{P}(s^*\lambda)$. Moreover, by similar analysis for $\nabla_\alpha S(\widetilde{\boldsymbol{\beta}}; \widehat{\mathbf{d}})$, it holds that

$$\nabla_\alpha S(\widetilde{\boldsymbol{\beta}}; \widehat{\mathbf{d}}) = \sigma^2(\boldsymbol{\beta}^*) + o_\mathbb{P}(1).$$

Together with (6.21), we conclude that term $(ii).b = o_\mathbb{P}(1)$. Plugging the above results into terms (i) and (ii), we finally have

$$\sqrt{n} \cdot [\bar{\alpha}(\widehat{\boldsymbol{\beta}}, \rho) - \alpha^*] \rightsquigarrow N(0, \sigma_0^2/\sigma^2(\boldsymbol{\beta}^*)). \tag{6.29}$$

As the final part of the proof, we show that $\sigma_W^2(\widehat{\boldsymbol{\beta}}, \rho)$ is a consistent estimator of $\sigma_0^2/\sigma^2(\boldsymbol{\beta}^*)$. Note that $\sigma_W^2(\widehat{\boldsymbol{\beta}}, \rho) = \widehat{\sigma}_0^2(\widehat{\boldsymbol{\beta}})/\widehat{\sigma}^2(\widehat{\boldsymbol{\beta}}, \rho)$ where $\widehat{\sigma}_0(\cdot)$ and $\widehat{\sigma}(\cdot, \cdot)$ are defined in (6.13) and (6.14) respectively. Similar to our analysis in §6.3, $\widehat{\sigma}_0^2(\widehat{\boldsymbol{\beta}})$ and $\widehat{\sigma}^2(\widehat{\boldsymbol{\beta}}, \rho)$ are consistent estimators of $\sigma_0^2$ and $\sigma^2(\boldsymbol{\beta}^*)$ individually. Since $\sigma^2(\boldsymbol{\beta}^*) > 0$, we conclude that $\sigma_W^2(\widehat{\boldsymbol{\beta}}, \rho)$ converges to $\sigma_0^2/\sigma^2(\boldsymbol{\beta}^*)$ in probability. Therefore, by Slutsky's theorem, it holds that

$$\sqrt{n} \cdot [\bar{\alpha}(\boldsymbol{\beta}, \rho) - \alpha^*]/\widehat{\sigma}_W(\widehat{\boldsymbol{\beta}}, \rho) \rightsquigarrow N(0, 1).$$

$\square$

# 7 Conclusion

We study parameter estimation and asymptotic inference for high dimensional regression under known nonlinear transform. We propose an $\ell_1$-regularized least-square estimator for estimation. Although the optimization problem is non-convex, we show that every stationary point converges to the true signal with the optimal statistical rate of convergence. In addition, we propose an efficient algorithm that successfully converges to a stationary point. Moreover, based on the stationary points, we propose the decorrelated score and Wald statistics to construct valid hypothesis tests and confidence intervals for the low-dimensional component of the high-dimensional parameter. Thorough numerical experiments are provided to back up the developed theory.

# A  Proof of Auxiliary Results

In this appendix, we provide the proofs of the auxiliary results. Specifically, we verify the conditions and prove the lemmas appearing in §6.

## A.1  Proof of Auxiliary Results for Estimation

In the first part of Appendix §A, we prove the auxiliary results used in the proof of Theorem (1). Meanwhile, we first verify the condition that is used to obtain the fast rates of convergence of the $\ell_1$-regularized estimator. In the proof we need to consider the concentration of terms involving $\{\mathbf{x}_i\}_{i=1}^n$, thus we introduce random vectors $\{\mathbf{x}_i\}_{i=1}^n$ that are i.i.d. copies of $\mathbf{x}$ and $\{\mathbf{x}_i\}_{i=1}^n$ are the realizations.

**Verify Condition** `Bounded-Gradient`$(B, \delta)$. By the definition of loss function $L$, for $j = 1, \ldots, d$, the $j$-th entry of $\nabla L(\boldsymbol{\beta}^*)$ can be written as $\nabla_j L(\boldsymbol{\beta}^*) = 1/n \cdot \sum_{i=1}^n \epsilon_i f'(\mathbf{x}_i^\top \boldsymbol{\beta}^*) x_{ij}$. Recall that $\epsilon_i$'s are i.i.d. centered sub-Gaussian random variables with variance proxy $\sigma^2$. Thus conditioning on $\mathbf{x}_1, \ldots, \mathbf{x}_n$, $\nabla_j L(\boldsymbol{\beta}^*)$ is a centered sub-Gaussian random variable with variance proxy bounded by

$$\sigma^2 \cdot \frac{1}{n^2} \sum_{i=1}^n f'(\mathbf{x}_i^\top \boldsymbol{\beta}^*)^2 x_{ij}^2 \leq \sigma^2 \cdot b^2 \cdot \widehat{\Sigma}_{j,j}/n, \quad \text{where } \widehat{\boldsymbol{\Sigma}} = \frac{1}{n} \sum_{i=1}^n \mathbf{x}_i \mathbf{x}_i^\top.$$

Under Assumption `Bounded-Design`$(D)$, the variance proxy of $\nabla_j L(\boldsymbol{\beta}^*)$ is bounded by $\sigma^2 \cdot b^2 \cdot D/n$. By the definition of variance proxy of sub-Gaussian random variables, we have

$$\mathbb{P}\big(\big|\nabla_i \mathcal{L}(\boldsymbol{\beta}^*)\big| > \sigma \cdot b \cdot t \cdot \sqrt{D/n} \big| \mathbf{x}_1, \ldots, \mathbf{x}_n\big) \leq 2\exp(-t^2/2), \quad \forall t > 0. \tag{A.1}$$

Taking a union bound over $j = 1, 2, \ldots, d$ in for the left-hand side of (A.1) we obtain that

$$\mathbb{P}\big(\big\|\nabla L(\boldsymbol{\beta}^*)\big\|_\infty > \sigma \cdot b \cdot t \sqrt{D/n} \big| \mathbf{x}_1, \ldots, \mathbf{x}_n\big) \leq 2\exp(-t^2/2 + \log d), \quad \forall t > 0. \tag{A.2}$$

By choosing $t = C\sqrt{\log d}$ in (A.2) for a sufficiently large $C$, we conclude that there exist a constant $B = C \cdot b \cdot \sqrt{D} > 0$ such that $\|\nabla L(\boldsymbol{\beta}^*)\|_\infty \leq B\sigma\sqrt{\log d/n}$ with probability at least $1 - \delta$, where we have $\delta \leq (2d)^{-1}$.

*Proof of Lemma 6.* By the definition of $L(\boldsymbol{\beta})$, the gradient $\nabla L(\boldsymbol{\beta})$ is given by

$$\nabla L(\boldsymbol{\beta}) = -\frac{1}{n} \sum_{i=1}^n [y_i - f(\mathbf{x}_i^\top \boldsymbol{\beta})] f'(\mathbf{x}_i^\top \boldsymbol{\beta}) \mathbf{x}_i. \tag{A.3}$$

Hence for $\nabla L(\boldsymbol{\beta}^*)$, (A.3) can be reduced to

$$\nabla L(\boldsymbol{\beta}^*) = -\frac{1}{n} \sum_{i=1}^n \epsilon_i f'(\mathbf{x}_i^\top \boldsymbol{\beta}^*) \mathbf{x}_i, \tag{A.4}$$

where $\epsilon_1, \ldots, \epsilon_n$ are $n$ i.i.d. realizations of the random noise $\epsilon$ in (1.1). For any $\boldsymbol{\beta} \in \mathbb{R}^d$, we denote $\boldsymbol{\eta} = \boldsymbol{\beta} - \boldsymbol{\beta}^*$. Recalling that $y_i = f(\mathbf{x}_i^\top \boldsymbol{\beta}^*) + \epsilon_i$, Taylor expansion of (A.3) implies that

$$\nabla L(\boldsymbol{\beta}) = -\frac{1}{n} \sum_{i=1}^n \epsilon_i f'(\mathbf{x}_i^\top \boldsymbol{\beta}) \mathbf{x}_i + \frac{1}{n} \sum_{i=1}^n f'(\mathbf{x}_i^\top \widetilde{\boldsymbol{\beta}}) f'(\mathbf{x}_i^\top \boldsymbol{\beta})(\mathbf{x}_i^\top \boldsymbol{\eta}) \mathbf{x}_i, \tag{A.5}$$

where $\widetilde{\boldsymbol{\beta}}$ lies on the line segment between $\boldsymbol{\beta}^*$ and $\boldsymbol{\beta}$. Combining (A.4) and (A.5) we have

$$\langle \nabla L(\boldsymbol{\beta}) - \nabla L(\boldsymbol{\beta}^*), \boldsymbol{\beta} - \boldsymbol{\beta}^* \rangle = A_1 + A_2, \tag{A.6}$$

where $A_1$ and $A_2$ are defined respectively as

$$A_1 = \frac{1}{n} \sum_{i=1}^n f'(\mathbf{x}_i^\top \widetilde{\boldsymbol{\beta}}) f'(\mathbf{x}_i^\top \boldsymbol{\beta})(\mathbf{x}_i^\top \boldsymbol{\eta})^2 \text{ and } A_2 = \frac{1}{n} \sum_{i=1}^n \{f'(\mathbf{x}_i^\top \boldsymbol{\beta}^*) - f'(\mathbf{x}_i^\top \boldsymbol{\beta})\}(\mathbf{x}_i^\top \boldsymbol{\eta}) \epsilon_i.$$

By the boundedness of $f'$, we can lower bound $A_1$ by

$$A_1 \geq a^2 \frac{1}{n} \sum_{i=1}^n (\mathbf{x}_i^\top \boldsymbol{\eta})^2 = a^2 \boldsymbol{\eta}^\top \widehat{\boldsymbol{\Sigma}} \boldsymbol{\eta}, \text{ where } \widehat{\boldsymbol{\Sigma}} := \frac{1}{n} \sum_{i=1}^n \mathbf{x}_i \mathbf{x}_i^\top. \tag{A.7}$$

For the second part $A_2$, by the sub-Gaussianity of $\epsilon_i$'s, $\{f'(\mathbf{x}_i^\top \boldsymbol{\beta}^*) - f'(\mathbf{x}_i^\top \boldsymbol{\beta})\}(\mathbf{x}_i^\top \boldsymbol{\eta})\epsilon_i$ is a centered sub-Gaussian random variable with variance proxy $\sigma^2[f'(\mathbf{x}_i^\top \boldsymbol{\beta}^*) - f'(\mathbf{x}_i^\top \boldsymbol{\beta})]^2 (\mathbf{x}_i^\top \boldsymbol{\eta})^2 \leq 4\sigma^2 b^2 (\mathbf{x}_i^\top \boldsymbol{\eta})^2$. Therefore we conclude that $A_2$ is centered and sub-Gaussian with variance proxy bounded by $4b^2 n^{-2} \sigma^2 \sum_{i=1}^n (\mathbf{x}_i^\top \boldsymbol{\eta})^2 = 4b^2 \sigma^2 n^{-1} \boldsymbol{\eta}^\top \widehat{\boldsymbol{\Sigma}} \boldsymbol{\eta}$. By the tail bound for sub-Gaussian random variables, we obtain that for any $x > 0$,

$$\mathbb{P}(|A_2| \geq x) \leq 2 \exp(-x^2/C), \text{ where } C = 8b^2 \sigma^2 n^{-1} \boldsymbol{\eta}^\top \widehat{\boldsymbol{\Sigma}} \boldsymbol{\eta}.$$

with probability at least $1 - (2d)^{-1}$, it holds that

$$A_2 \geq \sqrt{C \cdot \log(4d)} \geq -3b\sigma \sqrt{\log d/n} \sqrt{\boldsymbol{\eta}^\top \widehat{\boldsymbol{\Sigma}} \boldsymbol{\eta}} \geq -3b\sigma \sqrt{D \log d/n} \|\boldsymbol{\eta}\|_1, \tag{A.8}$$

where the last inequality is derived from Hölder's inequality $\boldsymbol{\eta}^\top \widehat{\boldsymbol{\Sigma}} \boldsymbol{\eta} \leq \|\widehat{\boldsymbol{\Sigma}}\|_\infty \|\boldsymbol{\eta}\|_1^2 \leq D \|\boldsymbol{\eta}\|_1^2$. Therefore combining (A.6), (A.7) and (A.8) with probability at least $1 - (2d)^{-1}$, we have

$$\langle \nabla L(\boldsymbol{\beta}) - \nabla L(\boldsymbol{\beta}^*), \boldsymbol{\beta} - \boldsymbol{\beta}^* \rangle \geq a^2 \boldsymbol{\eta}^\top \widehat{\boldsymbol{\Sigma}} \boldsymbol{\eta} - 3b\sigma \sqrt{D \log d/n} \|\boldsymbol{\eta}\|_1.$$

$\square$

*Proof of Lemma 7.* Recall that $\mathcal{J}$ is the set of indices of the largest $k^*$ entries of $\boldsymbol{\eta}_{\mathcal{S}^c}$ in absolute value and let $\mathcal{I} = \mathcal{J} \cup \mathcal{S}$. The following Lemma establishes a lower-bound on $\boldsymbol{\eta}^\top \widehat{\boldsymbol{\Sigma}} \boldsymbol{\eta}$. We prove this lemma in §A.3.

**Lemma 13.** Let $\boldsymbol{\Sigma} \in \mathbb{R}^{d \times d}$ be a positive semi-definite matrix and $\rho_-(k)$ and $\rho_+(k)$ be its $k$-sparse eigenvalues. Suppose that for some integer $s$ and $k$, $\rho_-(s + 2k) > 0$. For any $\mathbf{v} \in \mathbb{R}^d$, let $\mathcal{F}$ be any index set of size $d - s$, that is, $|\mathcal{F}^c| = s$. We let $\mathcal{J}$ be the set of indices of the $k$ largest component of $\mathbf{v}_{\mathcal{F}^c}$ in absolute value and let $\mathcal{I} = \mathcal{F}^c \cup \mathcal{J}$. Then we have

$$\mathbf{v}^\top \boldsymbol{\Sigma} \mathbf{v} \geq \rho_-(s + k) \cdot \left[ \|\mathbf{v}_\mathcal{I}\|_2 - \sqrt{\rho_+(k)/\rho_-(s+2k) - 1} \cdot \|\mathbf{v}_\mathcal{F}\|_1/\sqrt{k} \right] \|\mathbf{v}_\mathcal{I}\|_2.$$

By Assumption Sparse-Eigenvalue$(s^*, k^*)$, $\rho_-(s^* + 2k^*) > 0$. Combining Lemma 13 with $\mathcal{F} = \mathcal{S}^c$ and that $\|\boldsymbol{\eta}_{\mathcal{S}^c}\|_1 \leq \gamma \|\boldsymbol{\eta}_\mathcal{S}\|_1 \leq \gamma \sqrt{s} \|\boldsymbol{\eta}_\mathcal{S}\|_2$ together yield inequality (6.6).

For the second part of the lemma, by the definition of $\mathcal{J}$ we obtain that
$$\|\boldsymbol{\eta}_{\mathcal{I}^c}\|_\infty \leq \|\boldsymbol{\eta}_{\mathcal{J}}\|_1/k^* \leq \|\boldsymbol{\eta}_{\mathcal{S}^c}\|_1/k^* \leq \gamma/k^*\|\boldsymbol{\eta}_{\mathcal{S}}\|_1,$$
hence by Hölder's inequality we have
$$\|\boldsymbol{\eta}_{\mathcal{I}^c}\|_2 \leq \|\boldsymbol{\eta}_{\mathcal{I}^c}\|_1^{1/2}\|\boldsymbol{\eta}_{\mathcal{I}^c}\|_\infty^{1/2} \leq (\gamma/k^*)^{1/2}\|\boldsymbol{\eta}_{\mathcal{S}}\|_1^{1/2}\|\boldsymbol{\eta}_{\mathcal{I}^c}\|_1^{1/2} \leq \gamma k^{*-1/2} \cdot \|\boldsymbol{\eta}_{\mathcal{S}}\|_1,$$
where we use the fact that $\mathcal{I}^c \subset \mathcal{S}^c$. Thus it holds that
$$\|\boldsymbol{\eta}_{\mathcal{I}^c}\|_2 \leq \gamma\sqrt{s^*/k^*} \cdot \|\boldsymbol{\eta}_{\mathcal{S}}\|_2 \leq \gamma \cdot \|\boldsymbol{\eta}_{\mathcal{I}}\|_2 \text{ and } \|\boldsymbol{\eta}\|_2 \leq (1+\gamma) \cdot \|\boldsymbol{\eta}_{\mathcal{I}}\|_2. \tag{A.9}$$
□

## A.2 Proof of Auxiliary Results for Inference

We prove the auxiliary results for used to establish the validity of the decorrelated score and Wald test. As in §A.1, we start with verifying the conditions that are used in the theory of asymptotic inference.

**Verify Assumption `Bounded-Design` ($D$) for sub-Gaussian x.** We assume that $\mathbf{x}$ is i.i.d. sub-Gaussian. We denote $\boldsymbol{\Sigma} = \mathbb{E}\mathbf{x}\mathbf{x}^\top$. By the definition of $\widehat{\boldsymbol{\Sigma}}$, for any $j, k \in \{1, \ldots, d\}$, Cauchy-Schwarz inequality implies that

$$|\widehat{\Sigma}_{j,k}| = \Big|\frac{1}{n}\sum_{i=1}^n x_{ij}x_{ik}\Big| \leq \Big|\frac{1}{n}\sum_{i=1}^n x_{ij}^2\Big|^{1/2} \cdot \Big|\frac{1}{n}\sum_{i=1}^n x_{ik}^2\Big|^{1/2} = |\widehat{\Sigma}_{j,j}|^{1/2} \cdot |\widehat{\Sigma}_{k,k}|^{1/2}.$$

Hence we only need to bound the diagonal entries of $\widehat{\boldsymbol{\Sigma}}$. Note that $\mathbf{x}_1, \ldots, \mathbf{x}_n$ are i.i.d. centered sub-Gaussian random vectors. Hence for $j \in \{1, \ldots, d\}$, $X_{1j}^2, \ldots, X_{nj}^2$ are i.i.d. sub-exponential random variables (Vershynin, 2010). Applying Bernstein-type inequality to $\sum_{i=1}^n X_{ij}^2$, for any $t \geq 0$, we have

$$\mathbb{P}\Big\{\frac{1}{n}\sum_{i=1}^n [X_{ij}^2 - \mathbb{E}(X_{ij}^2)] > t\Big\} \leq 2\exp\big[-c \cdot \min(n \cdot t^2/K^2, \sqrt{n} \cdot t/K)\big], \tag{A.10}$$

where $c$ and $K$ are absolute constants that are uniformly for all $j$. Setting $t = C \cdot \sqrt{\log d/n}$ in (A.10) for some $c > 0$ and taking a union bound for $j = 1, \ldots, d$, we have $\|\widehat{\boldsymbol{\Sigma}} - \boldsymbol{\Sigma}\|_\infty \leq C \cdot \sqrt{\log d/n}$ with high probability. Therefore, when the sample size $n$ is sufficiently large, there exists a positive number $D > 0$ such that

$$\|\widehat{\boldsymbol{\Sigma}}\|_\infty \leq \|\boldsymbol{\Sigma}\|_\infty + C \cdot \sqrt{\log d/n} \leq D \text{ with probability at least } 1 - \delta,$$

where $\delta \leq (2d)^{-1}$.

**Verify Condition `Hessian-Stability`.** By definition, the Hessian of $L(\boldsymbol{\beta})$ is given by

$$\nabla^2 L(\boldsymbol{\beta}) = \frac{1}{n}\sum_{i=1}^n \big\{f'(\mathbf{x}_i^\top\boldsymbol{\beta})^2 - [y_i - f(\mathbf{x}_i^\top\boldsymbol{\beta})]f''(\mathbf{x}_i^\top\boldsymbol{\beta})\big\}\mathbf{x}_i\mathbf{x}_i^\top. \tag{A.11}$$

Note that $y_i = f(\mathbf{x}_i^\top\boldsymbol{\beta}^*) + \epsilon_i$, we have $\nabla^2 L(\boldsymbol{\beta}) - \nabla^2 L(\boldsymbol{\beta}^*) = \boldsymbol{A}_1 + \boldsymbol{A}_2 + \boldsymbol{A}_3$ where

$$\boldsymbol{A}_1 := \frac{1}{n}\sum_{i=1}^n [f'(\mathbf{x}_i^\top\boldsymbol{\beta})^2 - f'(\mathbf{x}_i^\top\boldsymbol{\beta}^*)^2]\mathbf{x}_i\mathbf{x}_i^\top, \quad \boldsymbol{A}_2 := \frac{1}{n}\sum_{i=1}^n \epsilon_i[f''(\mathbf{x}_i^\top\boldsymbol{\beta}) - f''(\mathbf{x}_i^\top\boldsymbol{\beta}^*)]\mathbf{x}_i\mathbf{x}_i^\top,$$

$$\text{and} \quad \boldsymbol{A}_3 := \frac{1}{n}\sum_{i=1}^n [f(\mathbf{x}_i^\top\boldsymbol{\beta}) - f(\mathbf{x}_i^\top\boldsymbol{\beta}^*)]f''(\mathbf{x}_i^\top\boldsymbol{\beta})\mathbf{x}_i\mathbf{x}_i^\top.$$

Under Assumption (3), for any $j, k \in \{1, \ldots, d\}$, by Taylor expansion, there exists a $\boldsymbol{\beta}_1$ in the line segment between $\boldsymbol{\beta}$ and $\boldsymbol{\beta}^*$ such that

$$\big|[\boldsymbol{A}_1]_{j,k}\big| \leq \frac{2}{n}\Big|\sum_{i=1}^n f'(\mathbf{x}_i^\top \boldsymbol{\beta}_1) f''(\mathbf{x}_i^\top \boldsymbol{\beta}_1) x_{ij} x_{ik} \mathbf{x}_i^\top (\boldsymbol{\beta} - \boldsymbol{\beta}^*)\Big| \leq a_{jk}^{(1)} \cdot \|\boldsymbol{\beta} - \boldsymbol{\beta}^*\|_1, \tag{A.12}$$

where we define $a_{jk}^{(1)} := \frac{2}{n}\Big\|\sum_{i=1}^n f'(\mathbf{x}_i^\top \boldsymbol{\beta}_1) f''(\mathbf{x}_i^\top \boldsymbol{\beta}_1) x_{ij} x_{ik} \mathbf{x}_i\Big\|_\infty.$

Here the second inequality of (A.12) follows from Hölder's inequality. Since $f'$ and $f''$ are both bounded, we immediately have $a_{jk}^{(1)} \leq 2bR \cdot \|n^{-1}\sum_{i=1}^n x_{ij} x_{ik} \mathbf{x}_i\|_\infty$.

Since $\mathbf{x}_1, \ldots, \mathbf{x}_n$ are i.i.d sub-Gaussian random vectors with variance proxy $\sigma_x^2$, for $1 \leq i \leq n$ and $1 \leq j \leq d$, we have $\mathbb{P}(|X_{ij} - \mathbb{E} X_j| > t) \leq 2\exp(-t^2/\sigma_x^2), \forall t \geq 0$. By taking a union bound for $i$ and $j$, there exist generic constants $C$ and $c$ such that event

$$\mathcal{A} := \{\|\mathbf{x}_i\|_\infty \leq C\sqrt{\log d}, 1 \leq i \leq n\} \tag{A.13}$$

holds with probability at least $1 - d^{-c}$. Conditioning on event $\mathcal{A}$, for $p \in \{1, \ldots, d\}$, Hoeffding's inequality implies that

$$\mathbb{P}\Big(\sum_{i=1}^n \big[X_{ij} X_{ik} X_{ip} - \mathbb{E}(X_{ij} X_{ik} X_{ip})\big] > t \Big| \mathcal{A}\Big) \leq 2\exp(-C \cdot t^2 \cdot n^{-1} \cdot \log^{-3} t), \quad \forall t > 0.$$

By taking a union bound over $j, k$ and $p$, we conclude that $a_{jk}^{(1)} \leq 2bR \cdot (C_1 + C_2 \log^2 d/\sqrt{n})$. By Assumption 4, it holds that $a_{jk}^{(1)}$'s are bounded uniformly for all $j, k$.

For $\boldsymbol{A}_2$, conditioning on $\{\mathbf{x}_i\}_{i=1}^n$, for any $j, k \in \{1, \ldots, d\}$, $[\boldsymbol{A}_2]_{j,k}$ is a centered sub-Gaussian random variable with variance proxy, denoted by $\mathrm{Var}_{jk}$, bounded by

$$\mathrm{Var}_{jk} \leq \frac{\sigma^2}{n^2} \sum_{i=1}^n [f''(\mathbf{x}_i^\top \boldsymbol{\beta}^*) - f''(\mathbf{x}_i^\top \boldsymbol{\beta})]^2 x_{ij}^2 x_{ik}^2.$$

Under Assumption 3, $\mathrm{Var}_{jk}$ can be further bounded by

$$\mathrm{Var}_{jk} \leq \frac{\sigma^2}{n^2} L_f^2 \sum_{i=1}^n x_{ij}^2 x_{ik}^2 (\boldsymbol{\beta} - \boldsymbol{\beta}^*)^\top \mathbf{x}_i \mathbf{x}_i^\top (\boldsymbol{\beta} - \boldsymbol{\beta}^*) \leq \frac{\sigma^2}{n} L_f^2 \|\boldsymbol{\beta} - \boldsymbol{\beta}^*\|_1^2 \cdot \|\boldsymbol{G}^{jk}\|_\infty, \tag{A.14}$$

where we denote $\boldsymbol{G}^{jk} := n^{-1}\sum_{i=1}^n x_{ij}^2 x_{ik}^2 \mathbf{x}_i \mathbf{x}_i^\top$. Conditioning on event $\mathcal{A}$, for any $p, q \in \{1, \ldots, d\}$, Hoeffding's inequality implies that

$$\mathbb{P}\big([\boldsymbol{G}^{jk}]_{p,q} - \mathbb{E}[\boldsymbol{G}^{jk}]_{p,q} > t \big| \mathcal{A}\big) \leq 2\exp(-C \cdot n \cdot t^2 \cdot \log^{-4} d), \quad \forall t > 0. \tag{A.15}$$

By setting $t = C \cdot \log^{5/2} d/\sqrt{n}$ and taking a union bound over $j, k, p$ and $q$, we obtain that $\|\boldsymbol{G}^{jk} - \mathbb{E}\boldsymbol{G}^{jk}\|_\infty \leq C \cdot \log^{5/2} d/\sqrt{n}$ with probability tending to one. Thus for all $j, k$, there exist a constant $G > 0$ such that $\|\boldsymbol{G}^{jk}\| \leq G + C \cdot \log^{5/2} d/\sqrt{n}$ with probability tending to one. Hence by (A.14), the variance proxy $\mathrm{Var}_{jk}$ is bounded by $\sigma^2/n \cdot L_f^2 \cdot \|\boldsymbol{\beta} - \boldsymbol{\beta}^*\|_1^2 \cdot (G + C \cdot \log^{5/2} d/\sqrt{n})$ with high probability. By Assumption 4, this implies that

$$\|\boldsymbol{A}_2\|_\infty = \mathcal{O}_{\mathbb{P}}\big[\|\boldsymbol{\beta} - \boldsymbol{\beta}^*\|_1 \cdot \sqrt{\log d/n} \cdot (G + C \cdot \log^{5/2} d/\sqrt{n})^{1/2}\big] = \mathcal{O}_{\mathbb{P}}(\|\boldsymbol{\beta} - \boldsymbol{\beta}^*\|_1).$$

Finally, for any $j, k \in \{1, \ldots, d\}$, by Taylor expansion and Hölder's inequality, we have $\big|[\boldsymbol{A}_3]_{j,k}\big| \leq bR \cdot \big\|n^{-1}\sum_{i=1}^n x_{ij} x_{ik} \mathbf{x}_i\big\|_\infty \cdot \|\boldsymbol{\beta} - \boldsymbol{\beta}^*\|_1$. By applying the technique for $[\boldsymbol{A}_1]_{j,k}$, we

conclude that $\|\boldsymbol{A}_3\|_\infty \leq \mathcal{O}_\mathbb{P}(\|\boldsymbol{\beta} - \boldsymbol{\beta}^*\|_1)$. Therefore, by combining the bounds for $\boldsymbol{A}_1, \boldsymbol{A}_2$ and $\boldsymbol{A}_3$, we conclude that $\|\nabla^2 L(\boldsymbol{\beta}) - \nabla^2 L(\boldsymbol{\beta}^*)\|_\infty = \mathcal{O}_\mathbb{P}(\|\boldsymbol{\beta} - \boldsymbol{\beta}^*\|_1)$.

**Verify Condition Hessian-Concentration($C^h$).** We denote $\boldsymbol{H} := n^{-1} \sum_{i=1}^n f'(\mathbf{x}_i^\top \boldsymbol{\beta}^*)^2 \cdot \mathbf{x}_i \mathbf{x}_i^\top$ and $\boldsymbol{E} := n^{-1} \sum_{i=1}^n \epsilon_i \cdot \mathbf{x}_i \mathbf{x}_i^\top$, then $\nabla^2 L(\boldsymbol{\beta}^*) = \boldsymbol{H} + \boldsymbol{E}$ and $\mathbf{I}(\boldsymbol{\beta}^*) = \mathbb{E}[\nabla^2 L(\boldsymbol{\beta}^*)] = \mathbb{E}[\boldsymbol{H}]$. Conditioning on $\{\mathbf{x}_i\}_{i=1}^n$, for any $j, k \in \{1, \ldots, d\}$, $E_{j,k}$ is a sub-Gaussian random variable with variance proxy bounded by $n^{-2} \sum_{i=1}^n x_{ij}^2 x_{ik}^2$, thus there exists a constant $C_1$ such that

$$\|\boldsymbol{E}\|_\infty \leq C_1 \sqrt{\log d/n} \cdot \max_{j,k} \Big(\frac{1}{n} \sum_{i=1}^n x_{ij}^2 x_{ik}^2\Big)^{1/2}. \tag{A.16}$$

with high probability. Similar to (A.15), we can also use Heoffding's inequality to conclude that $\|\mathbb{E}\|_\infty = \mathcal{O}_\mathbb{P}(\sqrt{\log d/n})$. For matrix $\boldsymbol{H}$, since $f'$ is bounded, $\{f'(\mathbf{x}_i^\top \boldsymbol{\beta}^*)\mathbf{x}_i, 1 \leq i \leq d\}$ are i.i.d. sub-Gaussian random vectors, hence each element of $f'(\mathbf{x}^\top \boldsymbol{\beta}^*)^2 \cdot \mathbf{x}\mathbf{x}^\top$ is a sub-exponential random variable with the same parameter. By applying the Bernstein-type inequality (A.10), we conclude that there exists a constant $C_2$ such that $\|\boldsymbol{H} - \mathbb{E}\boldsymbol{H}\|_\infty \leq C_2 \sqrt{\log d/n}$. This together with (A.16) implies Condition Hessian-Concentration($C^h$).

*Proof of Lemma 11.* We prove that $\|\mathbf{d}(\widehat{\boldsymbol{\beta}}, \rho) - \mathbf{d}^*\|_1 \leq 16/\tau_* \cdot \rho \cdot s_\mathbf{d}^*$. The proof is identical for $\mathbf{d}(\widehat{\boldsymbol{\beta}}', \rho)$. We first show that $\mathbf{d}^*$ defined in (3.7) is a feasible solution of the Dantzig selector problem defined in (2.7). By definition, we have

$$\begin{aligned}\left\|\nabla_{\boldsymbol{\gamma}\alpha} L(\widehat{\boldsymbol{\beta}}) - \nabla_{\boldsymbol{\gamma}\boldsymbol{\gamma}} L(\widehat{\boldsymbol{\beta}}) \cdot \mathbf{d}^*\right\|_\infty &= \left\|\nabla_{\boldsymbol{\gamma}\alpha} L(\widehat{\boldsymbol{\beta}}) - [\mathbf{I}(\boldsymbol{\beta}^*)]_{\boldsymbol{\gamma},\alpha} - \{\nabla_{\boldsymbol{\gamma}\boldsymbol{\gamma}} L(\widehat{\boldsymbol{\beta}}) - [\mathbf{I}(\boldsymbol{\beta}^*)]_{\boldsymbol{\gamma},\boldsymbol{\gamma}}\} \cdot \mathbf{d}^*\right\|_\infty \\ &\leq \|\nabla^2 L(\widehat{\boldsymbol{\beta}}) - \mathbf{I}(\boldsymbol{\beta}^*)\|_\infty \cdot (1 + \|\mathbf{d}^*\|_1). \end{aligned} \tag{A.17}$$

By triangle inequality, $\|\nabla^2 L(\widehat{\boldsymbol{\beta}}) - \mathbf{I}(\boldsymbol{\beta}^*)\|_\infty \leq \|\nabla^2 L(\widehat{\boldsymbol{\beta}}) - \nabla^2 L(\boldsymbol{\beta}^*)\|_\infty + \|\nabla^2 L(\boldsymbol{\beta}^*) - \mathbf{I}(\boldsymbol{\beta}^*)\|_\infty$. Condition Hessian-Stability implies that $\|\nabla^2 L(\widehat{\boldsymbol{\beta}}) - \nabla^2 L(\boldsymbol{\beta}^*)\|_\infty = \mathcal{O}_\mathbb{P}(\|\widehat{\boldsymbol{\beta}}' - \boldsymbol{\beta}^*\|_1)$. Condition Hessian-Concentration($C^h$) implies that $\|\nabla^2 L(\boldsymbol{\beta}^*) - \mathbf{I}(\boldsymbol{\beta}^*)\|_\infty \leq C^h \sqrt{\log d/n}$ with probability tending to one. By Theorem 1, we have $\|\widehat{\boldsymbol{\beta}}' - \boldsymbol{\beta}^*\|_1 = \mathcal{O}_\mathbb{P}(s^*\lambda)$ with $\lambda \asymp \sqrt{\log d/n}$. Hence from (A.17) we obtain that

$$\left\|\nabla_{\boldsymbol{\gamma}\alpha} L(\widehat{\boldsymbol{\beta}}) - \nabla_{\boldsymbol{\gamma}\boldsymbol{\gamma}} L(\widehat{\boldsymbol{\beta}}) \cdot \mathbf{d}^*\right\|_\infty \leq C \cdot s^* \lambda \cdot (1 + \|\mathbf{d}^*\|_1) \leq \rho,$$

which shows that $\mathbf{d}^*$ is a feasible solution of (2.7) with probability tending to one.

To obtain the statistical accuracy of $\mathbf{d}(\widehat{\boldsymbol{\beta}}, \rho)$, we define the restricted eigenvalue of $\nabla^2 L(\widehat{\boldsymbol{\beta}})$ on a index set $\mathcal{I}$ as

$$\widehat{\rho}(\mathcal{I}) := \inf_{\mathbf{v} \in \mathcal{C}} \frac{\mathbf{v}^\top \nabla^2 L(\widehat{\boldsymbol{\beta}}) \mathbf{v}}{\|\mathbf{v}\|_2^2}, \quad \text{where } \mathcal{C} = \{\mathbf{v} \in \mathbb{R}^d \colon \|\mathbf{v}_{\mathcal{I}^c}\|_1 \leq \|\mathbf{v}_\mathcal{I}\|_1, \mathbf{v} \neq \mathbf{0}\}. \tag{A.18}$$

The following lemma states that, for sufficiently large $n$, $\widehat{\rho}(\mathcal{I}) \geq \tau_*/2$ with high probability for any index set $\mathcal{I}$ satisfying $|\mathcal{I}| = s_\mathbf{d}^*$.

**Lemma 14.** *Under Assumptions 3 and 4, if the sample size $n$ is sufficiently large, for any index set $\mathcal{I}$ with $|\mathcal{I}| = s_\mathbf{d}^*$, we have $\widehat{\rho}(\mathcal{I}) \geq \tau_*/2 > 0$ with high probability.*

We denote $\mathcal{C} = \{\mathbf{v} \in \mathbb{R}^d \colon \|\mathbf{v}_{\mathcal{S}_\mathbf{d}^c}\|_1 \leq \|\mathbf{v}_{\mathcal{S}_\mathbf{d}}\|_1, \mathbf{v} \neq \mathbf{0}\}$ and from now on we condition on the event $\mathcal{E} := \{\widehat{\rho}(\mathcal{S}_\mathbf{d}) \geq \tau_*/2 > 0 \text{ and } \mathbf{d}^* \text{ is feasible}\}$. We denote $\boldsymbol{\omega} := \mathbf{d}(\widehat{\boldsymbol{\beta}}, \rho) - \mathbf{d}^*$ and $\widehat{\mathbf{d}} = \mathbf{d}(\widehat{\boldsymbol{\beta}}, \rho)$ for notational simplicity. The optimality of $\widehat{\mathbf{d}}$ implies that $\|\widehat{\mathbf{d}}\|_1 \leq \|\mathbf{d}^*\|_1$, which is equivalent to $\|\widehat{\mathbf{d}}_{\mathcal{S}_\mathbf{d}}\|_1 + \|\widehat{\mathbf{d}}_{\mathcal{S}_\mathbf{d}^c}\|_1 \leq \|\mathbf{d}_{\mathcal{S}_\mathbf{d}}^*\|_1$, where $\mathcal{S}_\mathbf{d}^c$ is the compliment of $\mathcal{S}_\mathbf{d}$. Triangle inequality implies that

$$\|\boldsymbol{\omega}_{\mathcal{S}_\mathbf{d}^c}\|_1 = \|\widehat{\mathbf{d}}_{\mathcal{S}_\mathbf{d}^c}\|_1 \leq \|\mathbf{d}_{\mathcal{S}_\mathbf{d}}^*\|_1 - \|\widehat{\mathbf{d}}_{\mathcal{S}_\mathbf{d}}\|_1 \leq \|\boldsymbol{\omega}_{\mathcal{S}_\mathbf{d}}\|_1, \quad \text{which implies that } (0, \boldsymbol{\omega}^\top)^\top \in \mathcal{C}.$$

Thus we have $\|\boldsymbol{\omega}\|_1 \leq 2\|\boldsymbol{\omega}_{\mathcal{S}_{\mathbf{d}}}\|_1 \leq 2\sqrt{s_{\mathbf{d}}^*} \cdot \|\boldsymbol{\omega}_{\mathcal{S}_{\mathbf{d}}}\|_2 \leq 2\sqrt{s_{\mathbf{d}}^*} \cdot \|\boldsymbol{\omega}\|_2$. Hence by Lemma 14, we have

$$\boldsymbol{\omega}^\top \nabla_{\gamma\gamma} L(\widehat{\boldsymbol{\beta}}) \, \boldsymbol{\omega} \geq \tau_*/2 \cdot \|\boldsymbol{\omega}\|_2^2 \geq \tau_*/(8 s_{\mathbf{d}}^*) \cdot \|\boldsymbol{\omega}\|_1^2. \tag{A.19}$$

Meanwhile, since both $\mathbf{d}^*$ and $\widehat{\mathbf{d}}$ are feasible, by triangle inequality we have

$$\|\nabla_{\gamma\gamma} L(\widehat{\boldsymbol{\beta}})\boldsymbol{\omega}\|_\infty \leq \|\nabla_{\gamma\alpha} L(\widehat{\boldsymbol{\beta}}) - \nabla_{\gamma\gamma} L(\widehat{\boldsymbol{\beta}})\widehat{\mathbf{d}}\|_\infty + \|\nabla_{\gamma\alpha} L(\widehat{\boldsymbol{\beta}}) - \nabla_{\gamma\gamma} L(\widehat{\boldsymbol{\beta}})\mathbf{d}^*\|_\infty \leq 2\rho.$$

Moreover, by Hölder's inequality, we obtain that

$$|\boldsymbol{\omega}^\top \nabla_{\gamma\gamma} L(\widehat{\boldsymbol{\beta}})\boldsymbol{\omega}| \leq \|\boldsymbol{\omega}\|_1 \|\nabla_{\gamma\gamma} L(\widehat{\boldsymbol{\beta}})\boldsymbol{\omega}\|_\infty \leq 2\rho \cdot \|\boldsymbol{\omega}\|_1. \tag{A.20}$$

Combining (A.19) and (A.20) we conclude that $\|\mathbf{d}(\widehat{\boldsymbol{\beta}}, \rho) - \mathbf{d}^*\|_1 = \|\boldsymbol{\omega}\|_1 \leq 16/\tau_* \cdot \rho \cdot s_{\mathbf{d}}^*$. Similarly, we also have, $\|\mathbf{d}(\widehat{\boldsymbol{\beta}}', \rho) - \mathbf{d}^*\|_1 = \|\boldsymbol{\omega}\|_1 \leq 16/\tau_* \cdot \rho \cdot s_{\mathbf{d}}^*$, which concludes the proof. $\square$

## A.3 Proof of Technical Lemmas

In the last part of Appendix §A, we prove two technical lemmas that are used in §A.1 and §A.2 respectively. The proof of Lemma 13 is inspired by Zhang (2010) and this lemma is potentially useful for other sparse estimation problems.

*Proof of Lemma 13.* Without loss of generality, we assume that $\mathcal{F}^c = \{1, \ldots, s\}$. We also assume that for $\mathbf{v} \in \mathbb{R}^d$, when $j > s$, $v_j$ is arranged in descending order of $|v_j|$. That is, we rearrange the components of $\mathbf{v}$ such that $|v_j| \geq |v_{j+1}|$ for all $j$ greater than $s$. Let $\mathcal{J}_0 = \{1, \ldots, s\}$ and $\mathcal{J}_i = \{s+(i-1)k+1, \ldots, \min(s+ik, d)\}$ for $i \geq 1$. By definition, we have $\mathcal{J} = \mathcal{J}_1$ and $\mathcal{I} = \mathcal{J}_0 \cup \mathcal{J}_1$. Moreover, we have $\|\mathbf{v}_{\mathcal{J}_i}\|_\infty \leq \|\mathbf{v}_{\mathcal{J}_{i-1}}\|_1/k$ when $i \geq 2$ because of the descending order of $|v_j|$ for $j > s$. Then we further have $\sum_{i \geq 2} \|\mathbf{v}_{\mathcal{J}_i}\|_\infty \leq \|\mathbf{v}_{\mathcal{F}}\|_1/k$.

We define the the restricted correlation coefficients of $\boldsymbol{\Sigma}$ as

$$\pi(s, k) := \sup\left\{\frac{\mathbf{v}_{\mathcal{I}}^\top \boldsymbol{\Sigma} \mathbf{v}_{\mathcal{J}} \|\mathbf{v}_{\mathcal{I}}\|_2}{\mathbf{v}_{\mathcal{I}}^\top \boldsymbol{\Sigma} \mathbf{v}_{\mathcal{I}} \|\mathbf{v}_{\mathcal{J}}\|_\infty} : I \cap \mathcal{J} = \emptyset, |\mathcal{I}| \leq s, |\mathcal{J}| \leq k, \mathbf{v} \in \mathbb{R}^d\right\}.$$

As shown in Zhang (2010), we have

$$\pi(s, k) \leq \frac{\sqrt{k}}{2} \cdot \sqrt{\rho_+(k)/\rho_-(s+k) - 1} \text{ if } \rho_-(s+k) > 0. \tag{A.21}$$

Then by the definition of $\pi(s+k, k)$ we obtain

$$\left|\mathbf{v}_{\mathcal{I}}^\top \boldsymbol{\Sigma} \mathbf{v}_{\mathcal{J}_i}\right| \leq \pi(s+k, k) \cdot \left(\mathbf{v}_{\mathcal{I}}^\top \boldsymbol{\Sigma} \mathbf{v}_{\mathcal{I}}\right) \cdot \|\mathbf{v}_{\mathcal{J}_i}\|_\infty / \|\mathbf{v}_{\mathcal{I}}\|_2.$$

Thus we have the following upper bound for $\left|\mathbf{v}_{\mathcal{I}}^\top \boldsymbol{\Sigma} \mathbf{v}_{\mathcal{I}^c}\right|$:

$$\begin{aligned}\left|\mathbf{v}_{\mathcal{I}}^\top \boldsymbol{\Sigma} \mathbf{v}_{\mathcal{I}^c}\right| &\leq \sum_{i \geq 2} \left|\mathbf{v}_{\mathcal{I}}^\top \boldsymbol{\Sigma} \mathbf{v}_{\mathcal{J}_i}\right| \leq \pi(s+k, k) \cdot \|\mathbf{v}_{\mathcal{I}}\|_2^{-1} \left(\mathbf{v}_{I}^\top \boldsymbol{\Sigma} \mathbf{v}_{\mathcal{I}}\right) \sum_{i \geq 2} \|\mathbf{v}_{\mathcal{J}_i}\|_\infty \\ &\leq \pi(s+k, k) \cdot \|\mathbf{v}_{\mathcal{I}}\|_2^{-1} \left(\mathbf{v}_{\mathcal{I}}^\top \boldsymbol{\Sigma} \mathbf{v}_{\mathcal{I}}\right) \|\mathbf{v}_{\mathcal{F}}\|_1/k. \end{aligned} \tag{A.22}$$

Because $\mathbf{v}^\top \boldsymbol{\Sigma} \mathbf{v} \geq \mathbf{v}_{\mathcal{I}}^\top \boldsymbol{\Sigma} \mathbf{v}_{\mathcal{I}} + 2\mathbf{v}_{\mathcal{I}}^\top \boldsymbol{\Sigma} \mathbf{v}_{\mathcal{I}^c}$, by (A.22) we have

$$\begin{aligned}\mathbf{v}^\top \boldsymbol{\Sigma} \mathbf{v} &\geq \mathbf{v}_{\mathcal{I}}^\top \boldsymbol{\Sigma} \mathbf{v}_{\mathcal{I}} - 2\pi(s+k, k) \|\mathbf{v}_I\|_2^{-1} \left(\mathbf{v}_I^\top \boldsymbol{\Sigma} \mathbf{v}_I\right) \|\mathbf{v}_{\mathcal{F}}\|_1/k \\ &= \left(\mathbf{v}_{\mathcal{I}}^\top \boldsymbol{\Sigma} \mathbf{v}_{\mathcal{I}}\right)\left[1 - 2\pi(s+k, k) \|\mathbf{v}_{\mathcal{I}}\|_2^{-1} \|\mathbf{v}_{\mathcal{F}}\|_1/k\right]. \end{aligned} \tag{A.23}$$

Combining (A.23), the fact that $\mathbf{v}_{\mathcal{I}}^\top \boldsymbol{\Sigma} \mathbf{v}_{\mathcal{I}} \geq \rho_-(s+k) \cdot \|\mathbf{v}_{\mathcal{I}}\|_2^2$ and (A.21) for $\pi(s+k, k)$, we conclude the proof of Lemma (13). $\square$

*Proof of Lemma 14.* By definition, given an index set $\mathcal{I}$ such that $|\mathcal{I}| \leq s_{\mathbf{d}}^*$, for any $\mathbf{v} \in \mathcal{C} = \{\mathbf{v} \in \mathbb{R}^d : \|\mathbf{v}_{\mathcal{I}^c}\|_1 \leq \|\mathbf{v}_{\mathcal{I}}\|_1, \mathbf{v} \neq \mathbf{0}\}$, we have

$$\widehat{\rho}(\mathcal{I}) \geq \inf_{\mathbf{v} \in \mathcal{C}} \{\mathbf{v}^\top \cdot \mathbf{I}(\boldsymbol{\beta}^*) \cdot \mathbf{v} - |\mathbf{v}^\top \cdot [\nabla^2 L(\widehat{\boldsymbol{\beta}}) - \mathbf{I}(\boldsymbol{\beta}^*)] \cdot \mathbf{v}|\} \cdot \|\mathbf{v}\|_2^{-2}. \tag{A.24}$$

By the definition of the restricted eigenvalue of $\mathbf{I}(\boldsymbol{\beta}^*)$ in (A.18), we have $\mathbf{v}^\top \cdot \mathbf{I}(\boldsymbol{\beta}^*) \cdot \mathbf{v} \geq \tau_* \|\mathbf{v}\|_2^2$ for all $\mathbf{v} \in \mathbb{R}^d$. By Hölder's inequality and (A.24) we have

$$\widehat{\rho}(\mathcal{I}) \geq \tau_* - \sup_{\mathbf{v} \in \mathcal{C}} \|\mathbf{v}\|_1^2 \cdot \|\mathbf{v}\|_2^{-2} \cdot \|\nabla^2 L(\widehat{\boldsymbol{\beta}}) - \mathbf{I}(\boldsymbol{\beta}^*)\|_\infty. \tag{A.25}$$

By the definition of $\mathcal{C}$, for any $\mathbf{v} \in \mathcal{C}$, $\|\mathbf{v}_{\mathcal{I}^c}\|_1 \leq \|\mathbf{v}_{\mathcal{I}}\|_1$. Thus we have

$$\|\mathbf{v}\|_1 \leq 2\|\mathbf{v}_{\mathcal{I}}\|_1 \leq 2\sqrt{|\mathcal{I}|} \cdot \|\mathbf{v}_{\mathcal{I}}\|_2 \leq 2\sqrt{s_{\mathbf{d}}^*} \cdot \|\mathbf{v}\|_2. \tag{A.26}$$

Therefore combining (A.25) and (A.26) we obtain that

$$\widehat{\rho}(\mathcal{I}) \geq \tau_* - 4 s_{\mathbf{d}}^* \cdot \|\nabla^2 L(\widehat{\boldsymbol{\beta}}) - \mathbf{I}(\boldsymbol{\beta}^*)\|_\infty. \tag{A.27}$$

Under Conditions Hessian-Stability and Hessian-Convergence($C^h$), it holds that

$$\|\nabla^2 L(\widehat{\boldsymbol{\beta}}) - \mathbf{I}(\boldsymbol{\beta}^*)\|_\infty \leq \|\nabla^2 L(\widehat{\boldsymbol{\beta}}) - \nabla^2 L(\boldsymbol{\beta}^*)\|_\infty + \|\nabla^2 L(\boldsymbol{\beta}^*) - \mathbf{I}(\boldsymbol{\beta}^*)\|_\infty = \mathcal{O}_\mathbb{P}(s^*\lambda). \tag{A.28}$$

Therefore, by (A.28) and Assumption (4) we obtain that

$$s_{\mathbf{d}}^* \cdot \|\nabla^2 L(\widehat{\boldsymbol{\beta}}) - \mathbf{I}(\boldsymbol{\beta}^*)\|_\infty = s_{\mathbf{d}}^* \cdot \mathcal{O}_\mathbb{P}(s^*\lambda) = \mathcal{O}_\mathbb{P}(s_{\mathbf{d}}^* \cdot \rho) = o(1),$$

which guarantees that $\widehat{\rho}(\mathcal{I}) \geq \tau_*/2$ when the sample size $n$ is sufficiently large. $\square$

*Proof of Lemma 12.* For $\boldsymbol{\Lambda}(\boldsymbol{\beta})$ defined in (6.13), triangle inequality implies

$$\|\boldsymbol{\Lambda}(\widehat{\boldsymbol{\beta}}') - \mathbb{E}[\boldsymbol{\Lambda}(\boldsymbol{\beta}^*)]\|_\infty \leq \underbrace{\|\boldsymbol{\Lambda}(\widehat{\boldsymbol{\beta}}') - \boldsymbol{\Lambda}(\boldsymbol{\beta}^*)\|_\infty}_{(i)} + \underbrace{\|\boldsymbol{\Lambda}(\boldsymbol{\beta}^*) - \mathbb{E}[\boldsymbol{\Lambda}(\boldsymbol{\beta}^*)]\|_\infty}_{(i)}. \tag{A.29}$$

We bound term (i) in (A.29) by direct computation. For any $j, k \in \{1, \ldots, d\}$, by mean-value theorem

$$[\boldsymbol{\Lambda}(\widehat{\boldsymbol{\beta}}')]_{j,k} - [\boldsymbol{\Lambda}(\boldsymbol{\beta}^*)]_{j,k} = \frac{2}{n} \sum_{i=1}^n f'(\mathbf{x}_i^\top \boldsymbol{\beta}_1) f''(\mathbf{x}_i^\top \boldsymbol{\beta}_1) x_{ij} x_{ik} \mathbf{x}_i^\top (\widehat{\boldsymbol{\beta}}' - \boldsymbol{\beta}^*), \tag{A.30}$$

where $\boldsymbol{\beta}_1$ is an intermediate value between $\widehat{\boldsymbol{\beta}}$ and $\boldsymbol{\beta}^*$. For notational simplicity, we denote $\boldsymbol{F}_i^{jk}(\boldsymbol{\beta}) \coloneqq 2 f'(\mathbf{x}_i^\top \boldsymbol{\beta}_1) f''(\mathbf{x}_i^\top \boldsymbol{\beta}_1) x_{ij} x_{ik} \mathbf{x}_i$. Then by definition,

$$[\boldsymbol{\Lambda}(\widehat{\boldsymbol{\beta}}')]_{j,k} - [\boldsymbol{\Lambda}(\boldsymbol{\beta}^*)]_{j,k} = \frac{1}{n} \sum_{i=1}^n \boldsymbol{F}_i^{jk}(\boldsymbol{\beta}_1)^\top \cdot (\widehat{\boldsymbol{\beta}}' - \boldsymbol{\beta}^*).$$

Note that $f'(x) \in [a, b]$ and that $|f''(x)| \leq R$ for any $x \in \mathbb{R}$. Similar to the technique of bounding term $[\boldsymbol{A}_1]_{j,k}$ in (A.12) in §A.2, we conclude that

$$\|\boldsymbol{\Lambda}(\widehat{\boldsymbol{\beta}}') - \boldsymbol{\Lambda}(\boldsymbol{\beta}^*)\|_\infty = \mathcal{O}_\mathbb{P}(\|\widehat{\boldsymbol{\beta}}' - \boldsymbol{\beta}^*\|_1) = \mathcal{O}_\mathbb{P}(s^*\lambda) \tag{A.31}$$

by Hölder's inequality.

Now we bound term (ii) in (A.29). Since $f'$ is bounded, $\{f'(\mathbf{x}_i^\top \boldsymbol{\beta}^*) \mathbf{x}_i, 1 \leq i \leq d\}$ are i.i.d. and sub-Gaussian. Thus the elements of $f'(\mathbf{x}^\top \boldsymbol{\beta}^*)^2 \cdot \mathbf{x}\mathbf{x}^\top$ are sub-exponential random variables with the same parameter. By applying the Bernstein-type inequality (A.10), it holds that

$$\|\boldsymbol{\Lambda}(\boldsymbol{\beta}^*) - \mathbb{E}[\boldsymbol{\Lambda}(\boldsymbol{\beta}^*)]\|_\infty = \mathcal{O}_\mathbb{P}(\sqrt{\log d/n}) = \mathcal{O}_\mathbb{P}(\lambda). \tag{A.32}$$

Therefore, combining (A.31) and (A.32), we conclude the proof of Lemma 12. $\square$

**Acknowledgements.** The research of Han Liu was supported by NSF CAREER Award DMS1454377, NSF IIS1408910, NSF IIS1332109, NIH R01MH102339, NIH R01GM083084, and NIH R01HG06841. The work of Y. Eldar was funded by the European Unions Horizon 2020 research and innovation programme under grant agreement ERC-BNYQ, by the Israel Science Foundation under Grant no. 335/14, and by ICore: the Israeli Excellence Center Circle of Light. Tong Zhang was supported by NSF IIS-1407939, NSF IIS-1250985, and NIH R01AI116744.